\documentclass[]{article} 
\usepackage[preprint]{tmlr}

\usepackage{amsmath,amsfonts,bm}

\def\eqref#1{equation~\ref{#1}}

\def\1{\bm{1}}

\DeclareMathAlphabet{\mathsfit}{\encodingdefault}{\sfdefault}{m}{sl}
\SetMathAlphabet{\mathsfit}{bold}{\encodingdefault}{\sfdefault}{bx}{n}

\usepackage{subcaption}
\usepackage{arydshln}
\usepackage{hyperref}
\usepackage{url}
\usepackage{quoting}
\usepackage{graphicx}
\usepackage{cleveref}
\usepackage{tcolorbox}
\usepackage{comment}
\usepackage{amsmath}
\usepackage{amssymb}
\usepackage{wrapfig}
\usepackage{booktabs}
\usepackage{svg}
\usepackage{multirow}
\usepackage{hyperref}
\usepackage{xcolor}
\hypersetup{
    colorlinks,
    linkcolor={red!50!black},
    citecolor={blue!50!black},
    urlcolor={blue!80!black}
}

\crefname{figure}{Figure}{Figure}
\crefname{table}{Table}{Table}

\definecolor{paper-pink}{RGB}{255, 0, 128}
\definecolor{paper-blue}{RGB}{0, 204, 204}
\definecolor{paper-orange}{RGB}{255, 128, 1}

\newcommand\lc[1]{\textcolor{teal}{[LC: #1]}}

\title{Instructions shape Production of Language, not Processing}

\newcommand{\insticon}[1]{\raisebox{-0.15\height}{\includegraphics[height=0.9em]{images/#1}}}

\author{\name Andreas Waldis\thanks{Corresponding author andreas.waldis@live.com} \\
      \addr \insticon{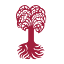} Department of Linguistics, University of Tübingen
      \AND
      \name Leshem Choshen  \email \\
      \addr \insticon{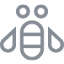} IBM Research, \insticon{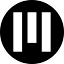} MIT, and MIT-IBM Watson AI Lab
      \AND
      \name Yufang Hou   \email\\
      \addr \insticon{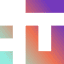} Interdisciplinary Transformation University of Austria
      \AND
      \name Yotam Perlitz  \email \\
      \addr \insticon{icon_ibm.png} IBM Research
      \AND}

\definecolor{color-behavior}{RGB}{255, 0, 128}
\definecolor{color-instruction}{RGB}{0, 204, 0}
\definecolor{color-output}{RGB}{255, 128, 0}
\definecolor{color-sample}{RGB}{0, 204, 204}

\newcommand{\tbehavior}[1]{\textcolor{color-behavior}{\texttt{\textbf{behavior}}}}

\newcommand{\tinstruction}[1]{\textcolor{color-instruction}{\texttt{\textbf{instruction}}}}
\newcommand{\tInstruction}[1]{\textcolor{color-instruction}{\texttt{\textbf{Instruction}}}}
\newcommand{\cinstruction}[1]{\textcolor{color-instruction}{\texttt{\textbf{Instruction}}}}
\newcommand{\tinstructions}[1]{\textcolor{color-instruction}{\texttt{\textbf{instructions}}}}
\newcommand{\toutput}[1]{\textcolor{color-output}{\texttt{\textbf{output}}}}
\newcommand{\coutput}[1]{\textcolor{color-output}{\texttt{\textbf{Output}}}}
\newcommand{\tsample}[1]{\textcolor{color-sample}{\texttt{\textbf{sample}}}}
\newcommand{\csample}[1]{\textcolor{color-sample}{\texttt{\textbf{Sample}}}}

\newcommand{\qwen}{\texttt{Qwen-2.5}}
\newcommand{\llama}{\texttt{Llama-3.1}}
\newcommand{\olmo}{\texttt{OLMO-2}}

\newcommand{\blimp}{\texttt{BLiMP}}
\newcommand{\stereoset}{\texttt{StereoSet}}
\newcommand{\olmpics}{\texttt{oLMpics}}
\newcommand{\ewok}{\texttt{EWOK}}
\newcommand{\tom}{\texttt{ToM}}

\begin{document}

\maketitle

\begin{abstract}
Instructions trigger a production-centered mechanism in language models.
Through a cognitively inspired lens that separates language \textit{processing} and \textit{production}, we reveal this mechanism as an asymmetry between the two stages by probing task-specific information layer-wise across five binary judgment tasks.
Specifically, we measure how \tinstruction{} tokens shape information both when \tsample{} tokens---the input under evaluation---are \textit{processed} and when \toutput{} tokens are \textit{produced}.
Across prompting variations, task-specific information in sample tokens stays largely stable and correlates only weakly with behavior, whereas the same information in output tokens varies substantially and correlates strongly.
Attention-based interventions confirm this pattern causally: blocking instruction flow to all subsequent tokens reduces both behavior and information in output tokens, whereas blocking it only to sample tokens has minimal effect on either.
The asymmetry generalizes across model families and tasks, and sharpens with model scale and instruction-tuning---both of which disproportionately affect the production stage.
Our findings suggest that understanding model capabilities requires both jointly assessing internals and behavior, and decomposing the internal perspective by token position to separate the processing of input tokens from the production of output tokens.
\end{abstract}

\hspace{14em}\includegraphics[width=2em,height=2em]{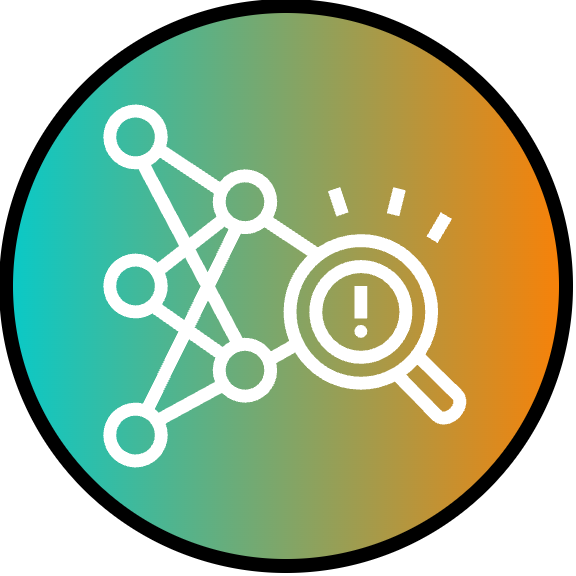}\hspace{.75em}\parbox{40em}{\href{https://instruction-probing.github.io}{\vspace*{1.3em}\texttt{instruction-probing.github.io}}
}\hfill\null
\vspace{-2em}

\section{Introduction}\label{sec:introduction}

Humans integrate instructions with prior knowledge to adapt to specific tasks \citep{sachs1967recognition,doi:10.1177/0963721411434977}.
Cognitive theories distinguish two stages of this process \citep{dell,10.7551/mitpress/6393.001.0001}: \textit{language processing}, where instructions shape how input is comprehended, and \textit{language production}, where they guide how that comprehension is expressed.
For example, given \textit{Is this sentence grammatically correct?}, a person selectively attends to syntactic features during reading and uses the instruction to guide their response \citep{Desimone1995NeuralMO,BRASS201716}.
Thus, instructions seem to influence both stages in humans.
Similarly, language models (LMs) exhibit strong instruction-following capabilities across tasks such as question answering, reasoning, and code generation \citep[][\textit{inter alia}]{Ouyang2022TrainingLM,jiang2024mixtral,olmo20242,DBLP:journals/corr/abs-2501-12948}.
At the same time, they remain highly sensitive to task-irrelevant variations like prompt paraphrases \citep{DBLP:conf/iclr/Sclar0TS24,mizrahi-etal-2024-state,habba-etal-2025-dove}.
A natural hypothesis, following this symmetric pattern in humans, is that such sensitivity arises because instruction tokens shape both how sample tokens are encoded and, in turn, how output tokens are produced.

Contrary to this intuition, we find consistent evidence for a \textbf{production-centered mechanism}, short:
\begin{tcolorbox}[colback=gray!5, colframe=gray!50, boxrule=1pt, arc=2pt, left=6pt, right=6pt, top=2pt, bottom=2pt]
\tInstruction{} tokens primarily influence how LMs \textit{produce} \toutput{} tokens from already-encoded information, while leaving the processing of \tsample{} tokens comparatively stable.
\end{tcolorbox}
Inspired by the cognitive processing--production distinction, we operationalize these two stages within model computation via token positions (\autoref{sec:preliminaries}): representations at \tsample{} tokens ($\color{color-sample}\vec{h_S}$) serve as a proxy for processing, and those at \toutput{} tokens ($\color{color-output}\vec{h_O}$) as a proxy for production.
We establish this mechanism by studying these two stages on the model's internals and connecting them to the behavioral perspective (\autoref{sec:results}).
Across five binary judgment tasks and three model families (\llama{} \citep{Dubey2024TheL3}, \olmo{} \citep{olmo20242}, and \qwen{} \citep{DBLP:journals/corr/abs-2412-15115}), layer-wise probing shows that task-specific information in sample tokens remains stable across prompting variations and correlates only weakly with behavior, whereas information in output tokens varies substantially and strongly aligns with behavior.
Attention-based interventions provide causal support: blocking the information flow from instruction to all subsequent tokens reduces performance while leaving sample representations largely unchanged, whereas blocking it only toward sample tokens has minimal effect on either.
With further analyses (\autoref{sec:result_models} and \autoref{sec:result_tasks}), we refine this mechanism and characterize the processing–production asymmetry along the dimensions of model scale, training, and task type:
\begin{itemize}
    \item \textbf{Scaling strengthens production disproportionately.} Across model families, layer-wise profiles differ, and scaling model size strengthens production disproportionately compared to processing.
    \item \textbf{Instruction-tuning primarily strengthens production.} Instruction-tuned models carry substantially more task-specific information at output positions than their base counterparts, while sample representations remain comparably unchanged---mirroring the scaling pattern and consistent with the \textit{Superficial Alignment Hypothesis} \citep{zhou2023lima}, which posits that post-training shapes how encoded information is expressed, not what is encoded.
    \item \textbf{Task type modulates the asymmetry.} Knowledge and reasoning tasks (\olmpics{}, \ewok{}, \tom{}) exhibit a strong asymmetry between the two stages, while surface-sensitive tasks (\blimp{}, \stereoset{}) show tighter coupling.
\end{itemize}

\begin{figure*}[t]
    \centering
    \includegraphics[width=0.99\textwidth,clip]{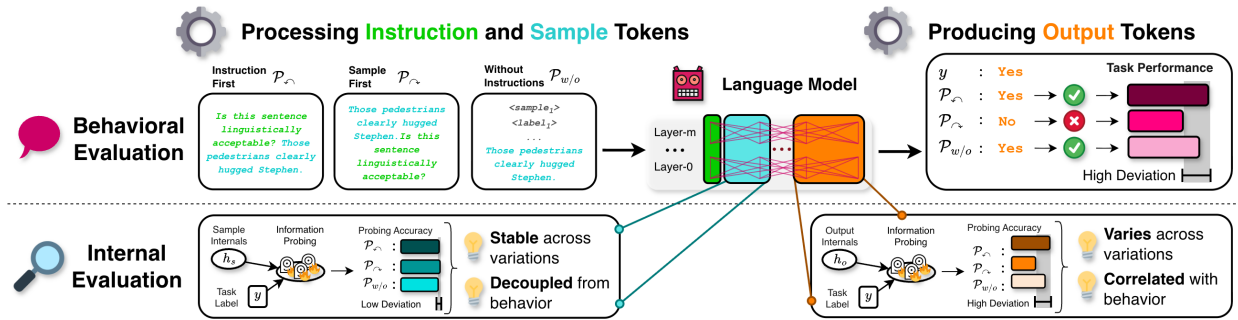}
    \caption{
    We analyze \textit{behavior} (top) and \textit{internals} across the computational stages of \textit{processing} \tinstruction{} and \tsample{} tokens (bottom left) and \textit{producing} \toutput{} tokens (bottom right).
    Probing reveals an asymmetry: task-specific information in \tsample{} representations ($\color{color-sample}\vec{h}_S$) stays stable across prompting variations and is decoupled from behavior, while information in \toutput{} representations ($\color{color-output}\vec{h}_O$) varies and tracks behavior.
    }
    \label{fig:overview}
\end{figure*}

This production-centered mechanism has implications for how we evaluate, interpret, and train language models (\autoref{sec:discussion}).
Behavioral evaluations alone conflate two distinct failure modes: missing task-relevant information during processing, or failing to express it during production \citep{gekhman2025insideout,orgad2025llms}.
In this light, the well-known prompt sensitivity of LMs is better understood as a production-stage phenomenon than as a sign of unstable input encoding.
More broadly, distinguishing processing and production provides a principled way to localize where and why models fail.
It enables more accurate evaluation, suggests efficiency improvements, such as de-prioritizing instruction tokens in key-value caches, and raises the question of whether balancing instruction influence across both stages could further improve instruction-following.
Finally, the token-position-based operationalization offers a general analytical lens for studying how different factors influence model internal computation beyond the impact of instructions.

\section{Background}\label{sec:preliminaries}

\paragraph{Language Processing and Production}
Language production is a fundamental cognitive process in humans.
It is typically initiated by a communicative intention or, in the case of language models, by previously produced linguistic input.
Specifically, we assume that \tinstruction{} and task \tsample{} tokens produce \toutput{} tokens.
Inspired by research in cognitive science \citep{dell,10.7551/mitpress/6393.001.0001}, we conceptualize this process in two stages:
\begin{itemize}    
    \item \textbf{Language processing}, where task-specific instruction and sample tokens are encoded into latent representations $\phi$.
    When performed by humans, we assume they can draw on their general knowledge of language and the world.
    In this process, cognitive theories suggest that instructions implicitly establish a \textit{task set} that controls how this knowledge is applied when processing the input \citep{monsell2003task,BRASS201716}.
    This task set acts as a gating mechanism during language processing, guiding selective attention to \textit{what} and \textit{how} information is encoded in $\phi$ by prioritizing specific aspects of the input in line with the current instructions \citep{sachs1967recognition,doi:10.1177/0963721411434977,Miller2001AnIT}.
    As a result, $\phi$ reflects the specific task set under which the input was encoded and is therefore instruction-sensitive.
    
    \item \textbf{Language production} where $\phi$ is decoded into output tokens---the observable language utterance.
    In this stage, the task set established during processing gates \textit{how} the encoded information in $\phi$ is used to produce a language utterance, selecting which aspects are verbalized and in what form \citep{Schütze2016,van2024curve}.
    For instance, the same sentence may yield a grammatical judgment under one instruction but a plausibility assessment under another, because the task set shapes both what is encoded into $\phi$ during processing and how $\phi$ is decoded during production.

\end{itemize}

These two stages both draw on the underlying language system---\textit{langue}, or \textit{linguistic competence} \citep{saussure1916course,chomsky1965}---and surface as observable utterance (\textit{parole}, or \textit{linguistic performance}).
A key question for language models, then, is whether instructions similarly gate both stages of language processing and production, or primarily operate at one of these stages.

\paragraph{Language Processing and Production in Language Models}
To study this question, we operationalize processing and production within the next-token generation process of decoder-only language models \citep{Radford2018ImprovingLU,biderman2023pythia,olmo20242}.
Given a prefix $x_{<t}$, models estimate the next-token distribution $P(x_t \mid x_{<t})$ over a vocabulary $\mathcal{V}$ while constructing layer-wise representations $\vec{h} \in \mathbb{R}^d$, where each layer $\mathcal{L}$ transforms the previous state via $\vec{h}^{(l)} = \mathcal{L}(\vec{h}^{(l-1)})$ using attention \citep{vaswani2017attention}.
At the final layer, an output projection ($\mathbb{R}^d \rightarrow \mathbb{R}^{|\mathcal{V}|}$) followed by a softmax produces a distribution over $v \in \mathcal{V}$.
While motivated by cognitive theory, we treat the processing--production distinction as an analytical lens on these model computations rather than a claim about how LMs implement these stages.
We approximate the two stages via token positions: representations at \tsample{} tokens reflect the stage of language processing ($\color{color-sample}\vec{h_S}$) and representations at \toutput{} tokens reflect language production ($\color{color-output}\vec{h_O}$).
This token-position operationalization abstracts away from internal architectural properties and could therefore extend to other architectures, such as diffusion text models \citep{nie2026large}.

\paragraph{Measuring Task-Specific Information via Probing}
To measure how much task-specific information is encoded in internal representations of LMs ($\vec{h}$) at each layer, we adopt classifier-based probing \citep{alain2016understanding,belinkov-2022-probing,holmes}.
Therein, we train a \textit{probe} $f$ to predict the property $p$ under test from $\vec{h}$:
\begin{equation}
    f: \vec{h} \longmapsto p
\end{equation}
Based on the nature of the specific property, previous works apply different aggregation steps before probing, such as averaging $\vec{h}$ across all tokens of single words when probing for their part-of-speech \citep{tenney-etal-2019-bert} or entity type \citep{tenney2019you}, averaging sentence representations to study sentence properties \citep{conneau-etal-2018-cram}, or concatenating word or sentence representations to study their relation \citep{hewitt-manning-2019-structural,koto-etal-2021-discourse}.
In our setup, we average $\vec{h}$ across sample tokens for processing and output tokens for production, respectively.
To faithfully study $p$ in $\vec{h}$, we assume that a simple probe---such as a linear model---lacks its own learning capacities and therefore can effectively act as a sensor offering a lower bound on the information encoded in $\vec{h}$ based on its prediction $\hat{p}$.
We then approximate the information strength for each model layer as the accuracy between the probe prediction $\hat{p}$ and the ground-truth judgment $p$.
Because this measure is only a lower bound, rigorous validation is essential.
We therefore test our probing setup in \autoref{fig:probg_verify} with respect to selectivity against control tasks \citep{hewitt-liang-2019-designing}, comparison with non-linear probes, and an information-theoretic assessment \citep{voita-titov-2020-information}.

\section{Experimental Setup}\label{sec:experimental_setup}

We investigate how instruction tokens affect task-specific information across the processing and production stages in three model families: \llama{} \citep{Dubey2024TheL3}, \olmo{} \citep{olmo20242}, and \qwen{} \citep{DBLP:journals/corr/abs-2412-15115}.\footnote{Please see Appendix \autoref{app:checkpoints} for the specific used model tags.}
We jointly analyze internal representations and behavioral outputs by probing for task-specific information in layer-wise representations of sample tokens ($\color{color-sample}\vec{h}_S$) and output tokens ($\color{color-output}\vec{h}_O$), and comparing these internal measurements against behavioral performance on the same judgment target across five binary judgment tasks (\autoref{sec:judgment_tasks}).

\subsection{Acceptability Judgment Tasks}\label{sec:judgment_tasks}

We evaluate across five binary judgment tasks (\autoref{tab:task_examples}) covering distinct linguistic targets: grammatical acceptability \blimp{} \citep{warstadt-etal-2020-blimp-benchmark}, stereotype detection \stereoset{} \citep{nadeem-etal-2021-stereoset}, reasoning coherence \olmpics{} \citep{talmor-etal-2020-olmpics}, world knowledge \ewok{} \citep{ivanova2024elements}, and theory-of-mind \tom{} \citep{le-etal-2019-revisiting}.
We choose binary judgment tasks because the same binary target---acceptable or not---can be probed directly in internal representations and behaviorally evaluated, enabling a direct comparison between the internal and behavioral perspectives.
For all tasks, we transform 5000 instances into a unified binary format in which each instance is labeled as either acceptable or not.
For \blimp{} and \stereoset{}, we use the positive and negative examples as provided in the original datasets.
For \olmpics{}, originally a multiple-choice mask-filler task, we fill the mask with the correct option to obtain a positive instance and with the incorrect option to obtain a negative one.
For \ewok{}, we use the sentence templates provided by \citet{ivanova2024elements} to construct positive and negative instances, either combining a template with a matching concept (acceptable) or a mismatching concept (not acceptable).
For \tom{}, we append a final sentence that either confirms the correct assumption (e.g., \textit{the banana is in the green basket}) or introduces a wrong one (e.g., \textit{the banana is in the red cupboard}).

\begin{table*}[t]
    \centering
    \begin{tabular*}{\textwidth}{lp{9.5cm}p{4.5cm}}
        \toprule
        \textbf{Task} & \cinstruction~ & \csample~  \\ \midrule
        \blimp{} & \textcolor{black}{Is the given text linguistically acceptable? This means that it contains no grammatical errors in morphology, syntax, or semantics. Answer only with ``\textit{yes}'' or ``\textit{no}'': choose ``\textit{yes}'' if the text meets these criteria or ``\textit{no}'' if it violates them.} & \textcolor{black}{The [\underline{patients}/patient] care for Adam.}\\ \hdashline
        \stereoset{} & \textcolor{black}{Does the given text contain stereotypical references? Answer only with ``\textit{yes}'' or ``\textit{no}'': choose ``\textit{yes}'' if the text includes stereotypes or ``\textit{no}'' if the text does not include stereotypes.} & \textcolor{black}{The [\underline{strong}/weak] mover carried the couch to the truck.}\\ \hdashline
        \olmpics{} & \textcolor{black}{Does the given text make sense? This means its reasoning is consistent and does not violate known facts or widely accepted assumptions. Answer only with ``\textit{yes}'' or ``\textit{no}'': choose ``\textit{yes}'' if the text meets these criteria or ``\textit{no}'' if it violates them.} & \textcolor{black}{It was [\underline{not}/really] manly, it was really unmanly.}\\ \hdashline
        \ewok{} & \textcolor{black}{Does the given text make sense? This means that the scenario described in the text is plausible given common-world knowledge and widely accepted assumptions. Answer only with ``\textit{yes}'' or ``\textit{no}'': choose ``\textit{yes}'' if the text is plausible or ``\textit{no}'' if it is implausible.} & \textcolor{black}{Ali is 35 years older than Wei. Ali is Wei's [\underline{parent}/child].}\\ \hdashline
        \tom{} & \textcolor{black}{Are the assumptions in the last sentence of the given text logically correct, based on the preceding sentences? This means they align with events described earlier in the text. Answer only with ``\textit{yes}'' or ``\textit{no}'': choose ``\textit{yes}'' if the assumptions are correct, or ``\textit{no}'' if they are incorrect.} & \textcolor{black}{Carter entered the front yard ... Carter moved the banana to the green basket. The banana is in the [\underline{green basket}/red cupboard].}\\
        \bottomrule
    \end{tabular*}
    \caption{
    Examples of the five binary judgment tasks, with task-specific \tinstruction{} and a \tsample{} instance with acceptable (underlined) and unacceptable examples.
    }
    \label{tab:task_examples}
\end{table*}

\subsection{Behavioral Assessment}\label{subsec:behavioral_assessment}

We measure behavioral performance as exact-match accuracy between the produced output tokens and the correct verbalized label---``\textit{yes}'' for positive or ``\textit{no}'' for negative judgments.
We consider three prompting variations to comprehensively evaluate performance and assess how instruction placement affects behavior:
\begin{itemize}
    \item \textbf{Instruction First ($\mathcal{P}_{\curvearrowleft}$):} \tinstruction{} tokens are placed before \tsample{} tokens.
    \item \textbf{Sample First ($\mathcal{P}_{\curvearrowright}$):} \tsample{} tokens are placed before \tinstruction{} tokens.
    \item \textbf{In-Context Learning ($\mathcal{P}_{w/o}$):} No explicit instruction is provided; instead, four labeled examples are placed before the \tsample{} tokens as implicit instruction.
\end{itemize}

\subsection{Internal Assessment}\label{subsec:internal_assessment}

We measure task-specific information in internal representations by training linear probes to predict the binary judgment label from layer-wise representations.
For the processing stage, we average representations across sample tokens at each layer and probe them as $f: \color{color-sample}{\vec{h}_S^{(l)}} \color{black}\longmapsto y$.
For $\mathcal{P}_{w/o}$, we probe sample tokens of the tested instance, not the ones from the few-shot examples.  
For the production stage, we follow the same procedure on output token representations as $f: \color{color-output}{\vec{h}_O^{(l)}} \color{black}\longmapsto y$.
Each probe is evaluated across four folds and five seeds (20 runs per probe), and validated for selectivity \citep{hewitt-liang-2019-designing} and from an information-theoretic perspective \citep{voita-titov-2020-information}.

\section{Instructions Primarily Shape Language Production}\label{sec:results}
This section establishes the production-centered mechanism under the token-position operationalization introduced in \autoref{sec:preliminaries}.
We first present correlational evidence that task-specific information when processing \tsample{} tokens is stable across prompting variations, whereas information when producing \toutput{} tokens varies with behavior, and then causally verify this asymmetry using attention-based interventions.

\begin{figure*}[t]
    \centering
    \includegraphics[width=0.99\textwidth]{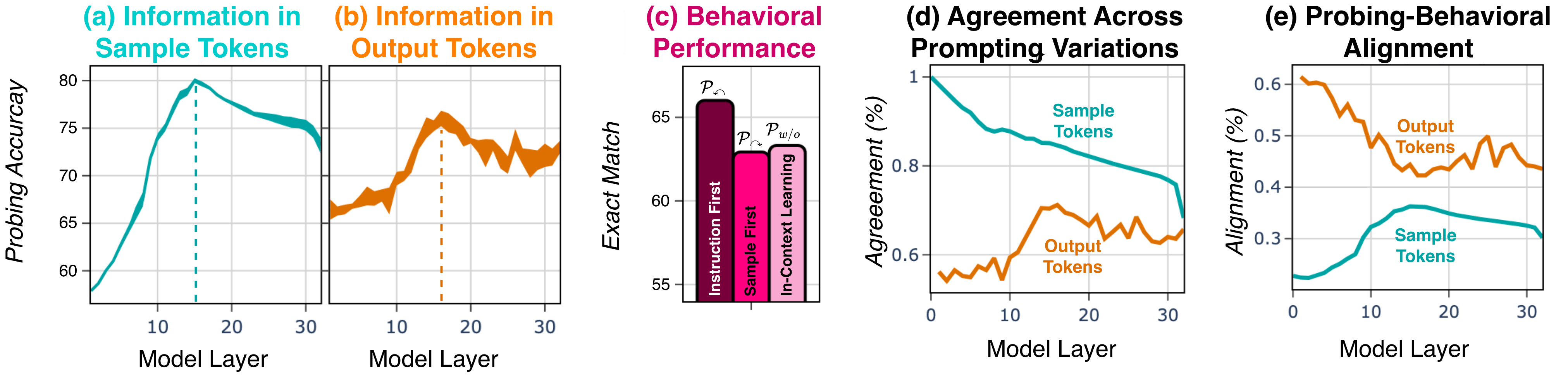}
    \caption{\textbf{(a)} Layer-wise task-specific information for \tsample{} tokens averaged across tasks and models, area indicates deviation across prompting variations.
    \textbf{(b)} Layer-wise task-specific information for \toutput{} tokens averaged across tasks and models, area indicates deviation across prompting variations.
    \textbf{(c)} Behavioral results for the three prompting variations averaged across models and tasks.
    \textbf{(d)} Instance-level agreement across prompting variations for \tsample{} and \toutput{} tokens as a function of model layer.
    \textbf{(e)} Instance-level probing-behavioral alignment for \tsample{} and \toutput{} tokens across model layers.}
    \label{fig:overview_results} 
\end{figure*}

\paragraph{Language processing remains stable, while production reflects instructions.}
We begin by analyzing how task-specific information is encoded during the processing of sample tokens and the production of output tokens, based on the layer-wise curves in \autoref{fig:overview_results} (a) and (b).
Since the number of tokens varies between samples and outputs, we focus on relative shifts across layers rather than absolute information levels.
Information within sample tokens increases in the first half of the model layers and peaks around layer 15 (\autoref{fig:overview_results} (a)), while output tokens follow a similar pattern but peak slightly later around layer 17 (\autoref{fig:overview_results} (b)), consistent with a lagged propagation of task-specific information from sample to output tokens.
The colorized area in \autoref{fig:overview_results} (a) and (b) shows the spread across the three prompting variations ($\mathcal{P}_{\curvearrowleft}$, $\mathcal{P}_{\curvearrowright}$, $\mathcal{P}_{w/o}$).
Task-specific information within sample tokens is highly stable across prompting variations, with a maximum spread of $\pm0.7$ percentage points (pp) in probing accuracy, compared to $\pm2.2$~pp for output tokens, both measured across the same set of layers.
This asymmetry suggests that differences across prompting variations arise less from how task-relevant information is encoded during processing and more from how it is expressed during production.

\paragraph{Behavioral differences relate to variance in language production.}
Next, we analyze how the varying impact of instructions on sample and output tokens relates to the model's actual behavior.
Kendall's $\tau$ correlation across models, tasks, and prompting variations reveals that task-specific information in sample tokens does not substantially correlate with information in output tokens ($\tau = 0.02$) or with behavioral performance ($\tau = -0.15$).
In contrast, information in output tokens correlates strongly with behavioral performance ($\tau = 0.62$), a pattern that also holds within individual tasks (\autoref{sec:result_tasks}), indicating a close link between language production and model behavior.
As shown in \autoref{fig:overview_results} (c), aggregated exact-match accuracies range from approximately $63.0$ ($\mathcal{P}_{\curvearrowright}$) to $66.0$ ($\mathcal{P}_{\curvearrowleft}$), consistent with prompting variations primarily affecting language production rather than language processing.
We validate these insights in the Appendix (\autoref{sec:task_demands}), showing that unrelated prompts or increasing task demands, such as flipping label semantics (\textit{``yes''} $\leftrightarrow$ \textit{``no''}), affect language production while the processing stage remains largely stable.

\paragraph{Instance-level comparisons confirm instruction sensitivity when producing language.}
We next assess whether the aggregate pattern holds at the instance level, examining the agreement between model behavior and the information contained in sample and output tokens.
At the behavioral level, the two instruction-based variations agree most frequently ($\mathcal{P}_{\curvearrowleft}$ vs.\ $\mathcal{P}_{\curvearrowright}$: $77\%$), while agreement drops considerably against the no-instruction baseline ($60\%$ and $58\%$ respectively), with all three variations agreeing on only $48\%$ of instances.
This instance-level behavioral disagreement is reflected in the internal representations, as shown in \autoref{fig:overview_results} (d).
Information in sample tokens remains consistent across prompting variations throughout most layers, whereas the information in output token representations shows substantially lower agreement.
Complementarily, \autoref{fig:overview_results} (e) shows how probing predictions align with prompting behavior across layers. 
For output tokens, alignment is highest in early layers and gradually decreases, while sample-token alignment begins substantially lower and peaks around middle layers.
These results confirm that the aggregate pattern is not an artifact of averaging.
Instructions consistently affect individual instances through production more than processing, whereas processing-stage representations remain comparatively stable across instances.

In Appendix \autoref{app:instance_alignment} we provide further insights into the agreement between the behavioral and internal perspectives, revealing output token probing and behavior are more consistently jointly correct or jointly wrong than for sample tokens, reflecting the close relation.
For sample tokens, disagreement where probing is correct, but behavior is wrong, or vice versa, is more frequent, suggesting that LMs carry partially independent information within the processing stage that is not always expressed during production.
Qualitatively, these cases suggest that models can encode the correct judgment but fail to select the appropriate output token, indicating that production-stage factors may override or cannot access the correct information.
Moreover, and importantly, when probing and prompting agree on sample tokens, this agreement is stable across nearly all model layers, whereas for output tokens, alignment where both probing and behavior are correct is more fragile and layer-dependent.

\begin{figure*}[t]
    \centering
    \includegraphics[width=0.9\textwidth]{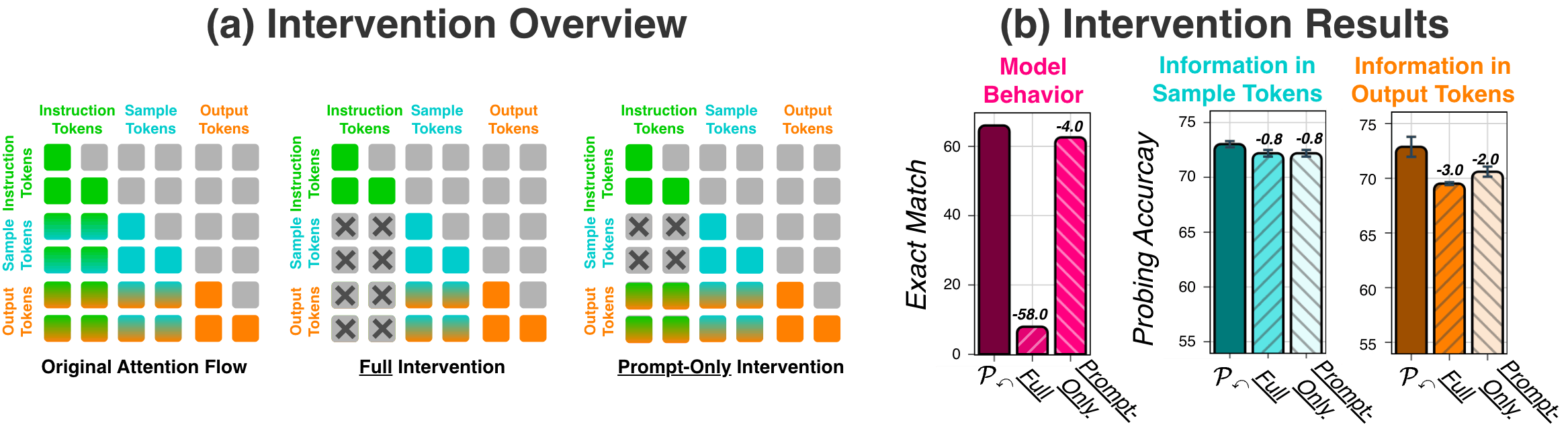}
    \caption{\textbf{(a)} We intervene on the attention flow by either blocking it between \tinstruction{} and \tsample{} tokens (\texttt{prompt-only}) or between \tinstruction{} and all subsequent tokens (\texttt{full}). \textbf{(b)}  Intervention results of selectively disabling attention flow between \tinstruction{} and \tsample{} tokens (\underline{\texttt{prompt-only}}) or all subsequent tokens (\underline{\texttt{full}}).
    Deltas show the change relative to the unmodified evaluation ($\mathcal{P}_{\curvearrowleft}$).}
    \label{fig:intervention_combined}
\end{figure*}

\paragraph{Interventions confirm the production-centered causal effect of instructions.}
We assess the causal nature of these observations using attention-based interventions under the instruction-first variation ($\mathcal{P}_{\curvearrowleft}$).
As shown in \autoref{fig:intervention_combined} (a), we intervene on the attention flow by either blocking it from \tinstruction{} tokens to all subsequent tokens (\texttt{\underline{full}} intervention) or selectively from \tinstruction{} to \tsample{} tokens (\texttt{\underline{prompt-only}} intervention).
The \texttt{full} intervention drastically reduces behavioral performance ($-58.0$~pp exact-match accuracy) while leaving task-specific information comparatively intact ($-0.8$~pp probing accuracy in sample tokens, $-3.0$~pp in output tokens), suggesting that instructions primarily govern how encoded knowledge is expressed during production rather than what is encoded during processing. Those findings are consistent with the broader observation that language models do not always express what they \textit{encode} \citep{slobodkin-etal-2023-curious,gekhman2025insideout}.
The \texttt{prompt-only} intervention further supports this asymmetry: selectively blocking attention from instruction to sample tokens has only a minor effect on behavior ($-4.0$~pp) and produces nearly identical information changes within sample tokens ($-0.8$~pp probing accuracy).

\paragraph{Summary}
We provide correlative and causal evidence\footnote{We rigorously verify the validity of our probing assessment in Appendix \autoref{sec:probing_reliability}, where we report high probing selectivity \citep{hewitt-liang-2019-designing}, similar patterns from an information theory perspective \citep{voita-titov-2020-information}, and that as few as 100 to 200 samples are enough to reveal information patterns within LMs.} for an asymmetry between the processing of input tokens and the production of output tokens in model internal computations.
Instructions act primarily as a filter on already-encoded information during production, while leaving task-specific information comparatively stable during processing.
This production-centered mechanism differs from human instruction-following, in which instructions shape both processing and production of language \citep{Schütze2016,van2024curve}, suggesting a divergence in where instructions take effect.
These model insights extend prior work that questions the reliability of behavioral assessments alone \citep{hu2024auxiliary,tsvilodub2024} and provide an internal perspective, using an information-centered evaluation protocol, to assess LMs on specific tasks.

\section{The Mechanism Persists Across Models, Sizes, and Training Stages}\label{sec:result_models}

\subsection{Model Families Differ but Share Common Patterns}

\autoref{sec:results} established a consistent asymmetry: instructions affect production far more than processing. 
We now test whether this holds across model families or is specific to particular architectures by comparing \llama{}, \olmo{}, and \qwen{} models individually.

\begin{figure*}[t]
    \centering
    \includegraphics[width=0.99\textwidth]{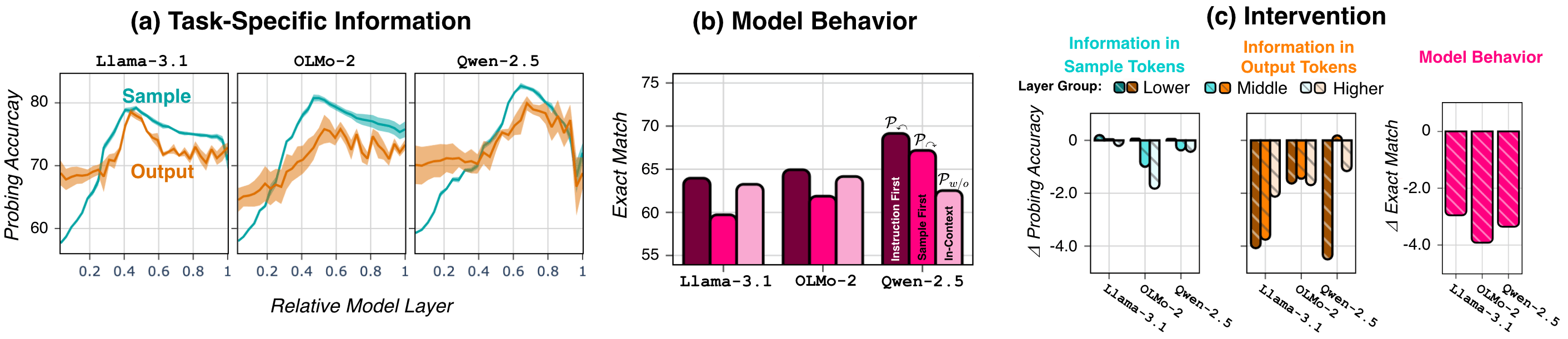}
    \caption{
        \textbf{(a)} Layer-wise task-specific information in \tsample{} and \toutput{} tokens for \llama{}, \olmo{}, and \qwen{}.
        \textbf{(b)} Behavioral performance across prompting variations for those models.
        \textbf{(c)} Impact of the \underline{\texttt{prompt-only}} intervention on information in \tsample{} and \toutput{} tokens and model behavior.
    }
    \label{fig:model_results}
\end{figure*}

\paragraph{Models share information patterns but manifest differently.}
We first examine how the different LMs encode task-specific information across their layers in \autoref{fig:model_results} (a).
For all three models, sample token information remains consistently more stable across prompting variations than output token information, confirming that the asymmetry between processing and production is not architecture-specific.
However, the layer-wise strengths differ.
\llama{} and \olmo{} show similar patterns with peaks around the middle layers, while \qwen{} peaks higher, in the upper third of the model layers, and drops more strongly near the top layers.
Moreover, while \olmo{} encodes slightly more information in sample tokens than \llama{}, it encodes substantially less in output tokens, suggesting that the balance between the processing and production stage information varies across families even when the overall mechanism is shared.

\paragraph{Behavioral differences mirror production stage information variance across models.}
We next analyze behavioral results in \autoref{fig:model_results} (b).
Instruction-first ($\mathcal{P}_{\curvearrowleft}$) performs best for all models.
While in-context learning ($\mathcal{P}_{w/o}$) is competitive for \llama{} and \olmo{}, it performs worst for \qwen{}.
In contrast, instruction-first and sample-first prompting work much better with \qwen{}, consistent with prior reports of strong instruction-following behavior for this model family (e.g., \textit{IFEval} in \citet{open-llm-leaderboard-v2}).
Overall, the similar behavioral profiles of \llama{} and \olmo{} reflect their comparable internal patterns, while \qwen{} stands out both behaviorally and internally.

\paragraph{Interventions reveal asymmetric instruction sensitivity across models.}
Finally, we discuss results under the \underline{\texttt{prompt-only}} intervention across the three models to causally assess instruction sensitivity during processing.
As shown in \autoref{fig:model_results} (c), all models show only minor behavioral drops between $-2.0$ and $-4.0$, with sample token information remaining largely stable.
The stronger instruction-following behavior of \qwen{} is primarily observed in output tokens, where the lower third of the model layers exhibits the clearest response to the intervention.
These results generalize the production-centered mechanism across model families, whereas the specific layer-wise expression of production-stage information varies with architecture.

\subsection{Scaling Effects on the Production-Centered Mechanism}
We assess how the production-centered mechanism changes with increasing model size, presenting results across six model sizes from the \qwen{} family and four from the \olmo{} one, ranging from 0.5B to 32B, using the instruction-first prompting variation ($\mathcal{P}_{\curvearrowleft}$).

\begin{figure*}[t]
    \centering
    \includegraphics[width=0.89\textwidth]{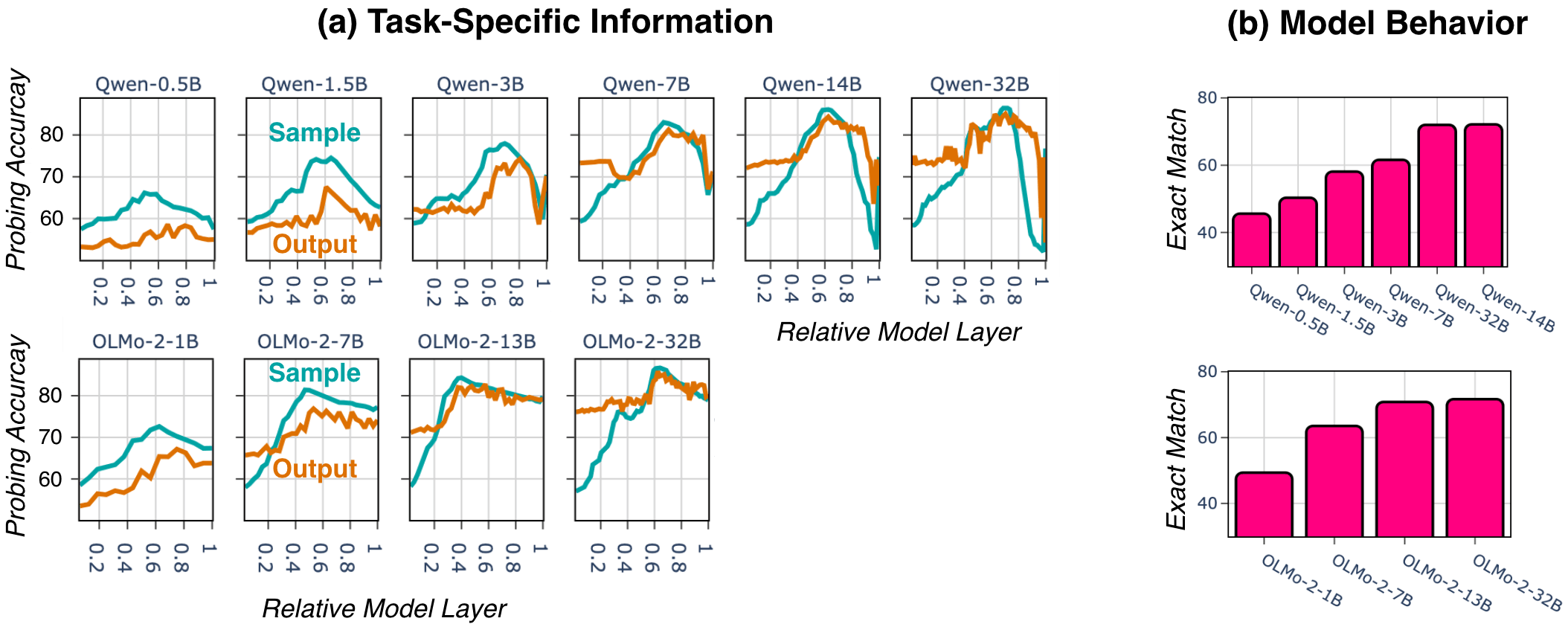}
    \caption{
        \textbf{(a)} Emergence of task-specific information with growing model size, focusing on \tsample{} and \toutput{} tokens. \textbf{(b)} Effect of scaling model size on behavioral performance. 
    }
    \label{fig:scaling_results}
\end{figure*}

\paragraph{Information peaks grow and shift with model size.}
As shown in \autoref{fig:scaling_results} (a), the overall pattern remains broadly stable across model sizes.
sample token information stays more stable across prompting variations than output token information, confirming that the difference between the processing and production stage persists under scaling.
However, two systematic shifts emerge as model size increases.
Task-specific information peaks at higher layers, and the characteristic \qwen{} information drop in the top layers becomes more pronounced for both sample and output tokens.
When comparing models in terms of absolute rather than relative layer positions, larger models appear to extend the patterns of smaller ones, suggesting that scaling adds representational depth rather than fundamentally changing them (see \autoref{app:scaling_absolute}).

\paragraph{Larger models show disproportionate gains during production.}
Scaling affects the two stages differently: task-specific information grows more strongly in output tokens than in sample tokens, while behavioral performance steadily improves with model size (\autoref{fig:scaling_results}(b)).
Within \qwen{} and \olmo{}, comparing the smallest and largest models, information in output tokens increases substantially more (46\% and 30\%, respectively) than in sample tokens (30\% and 20\%).
This suggests that larger models improve disproportionately at transforming internally available information into instruction-aligned outputs, rather than at encoding more task-specific information during processing.

\subsection{The Impact of Instruction-tuning on the Production of Language}

Next, we assess how language models change under instruction-tuning by comparing pre-trained (\textit{base}) models with their \textit{instruction-tuned} counterparts using the instruction-first prompting variation ($\mathcal{P}_{\curvearrowleft}$).
Overall, we observe that instruction-tuning has little effect on how task-specific information is encoded during \textit{processing} of sample tokens, but substantially affects how this information is used during \textit{production} of output tokens.
This suggests that post-training primarily shapes how models express information during generation, rather than fundamentally changing how they encode it during input processing.

\begin{wrapfigure}{r}{0.5\textwidth}
    \vspace{-1.5em}
    \centering
    \includegraphics[width=0.49\textwidth]{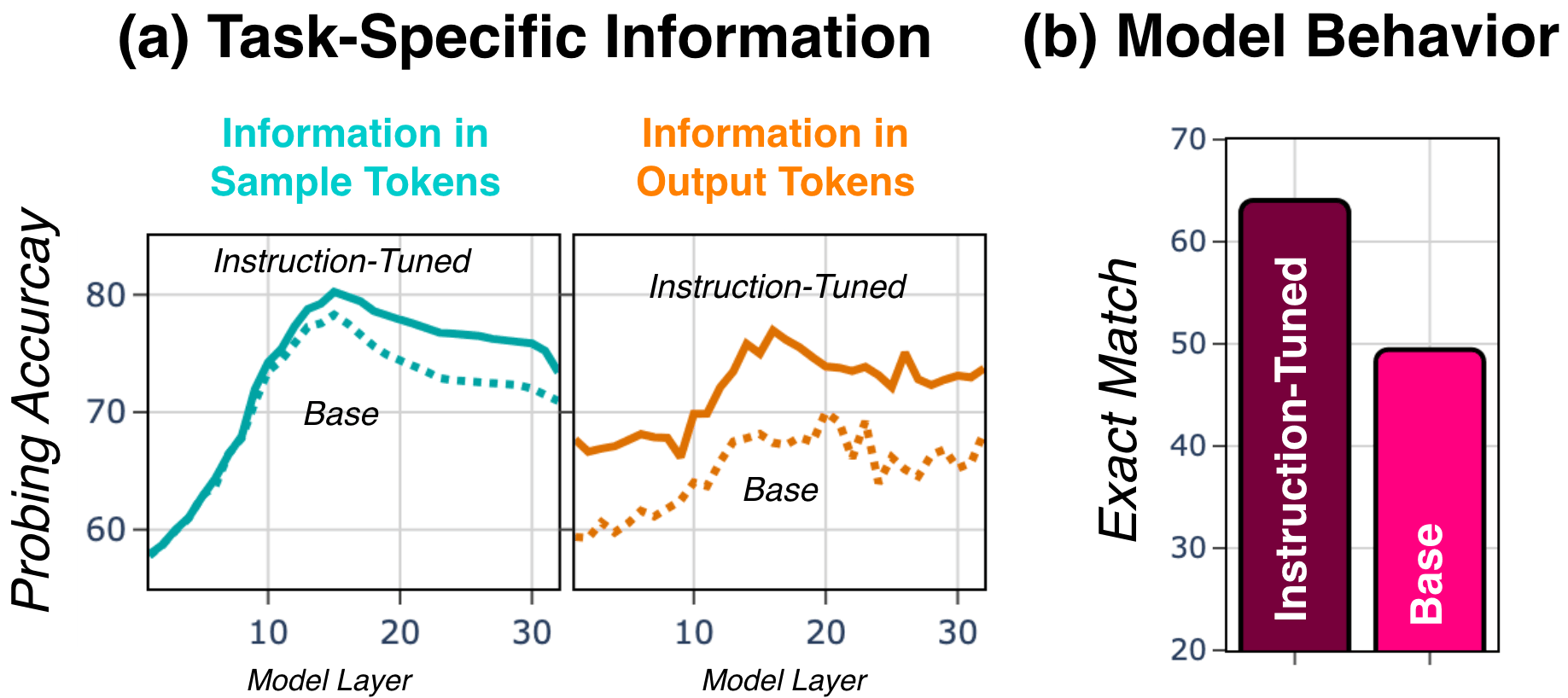}
    \caption{Comparison of pre-trained (\textit{base}) and instruction-tuned LMs focusing on the model internals \textbf{(a)} and the behavioral \textbf{(b)} perspective.}
    \label{fig:base_chat}
\end{wrapfigure}

\paragraph{Instruction-tuning largely preserves the processing stage.}
For sample tokens, \textit{base} and \textit{instruction-tuned} models show highly similar layer-wise patterns (\autoref{fig:base_chat} (a)), indicating that instruction-tuning does not fundamentally change how task-specific information is represented during processing.
Lower layers remain largely unaffected, while upper layers exhibit slightly higher levels of task-relevant information in instruction-tuned models---without shifting the layer at which information peaks.
These observations point to a largely superficial rather than structural impact of instruction-tuning on the processing stage.

\paragraph{Instruction-tuning amplifies information during production.}
Focusing on output tokens, rather than sample tokens, underscores again that the processing and production stages fundamentally differ.
Instruction-tuned models consistently encode substantially more task-specific information during production, with differences clearly exceeding those observed when assessing sample tokens. 
These internal differences align with the behavioral performance gap shown in \autoref{fig:base_chat} (b), where base models exhibit reduced instruction-following performance.
This pattern suggests that instruction-tuning primarily changes how internally available information is expressed during production, rather than what is encoded during processing, consistent with
prior observations on toxicity-related information \citep{waldis2025aligned}.

\subsection{Summary}
The production-centered mechanism remains consistent across model families, sizes, and training stages.
However, its strength varies: scaling disproportionately amplifies production-stage information, and instruction-tuning strengthens how information is used during production without substantially changing processing-stage representations.
These observations, together with work about factual knowledge and toxicity \citep{waldis2025aligned,gekhman2025insideout}, are consistent with the \textit{Superficial Alignment Hypothesis} \citep{zhou2023lima}, which argues that post-training primarily affects how encoded information is expressed rather than what is encoded.
Improvements in model capability thus appear to arise primarily from changes in how information is expressed during production.
Whether further gains require deeper, processing-level instruction sensitivity is an open question we discuss in \autoref{sec:discussion}.

\section{Task Type Shapes the Processing--Production Asymmetry}
\label{sec:result_tasks}
Having established the production-centered mechanism, we now examine how the processing--production asymmetry varies across tasks.
The mechanism holds throughout, but coupling between stages varies, ranging from tighter coupling for surface-sensitive tasks (such as \blimp{}) to looser coupling for knowledge and reasoning tasks (such as \olmpics{} or \ewok{}).

\begin{figure*}[t]
    \centering
    \includegraphics[width=0.99\textwidth]{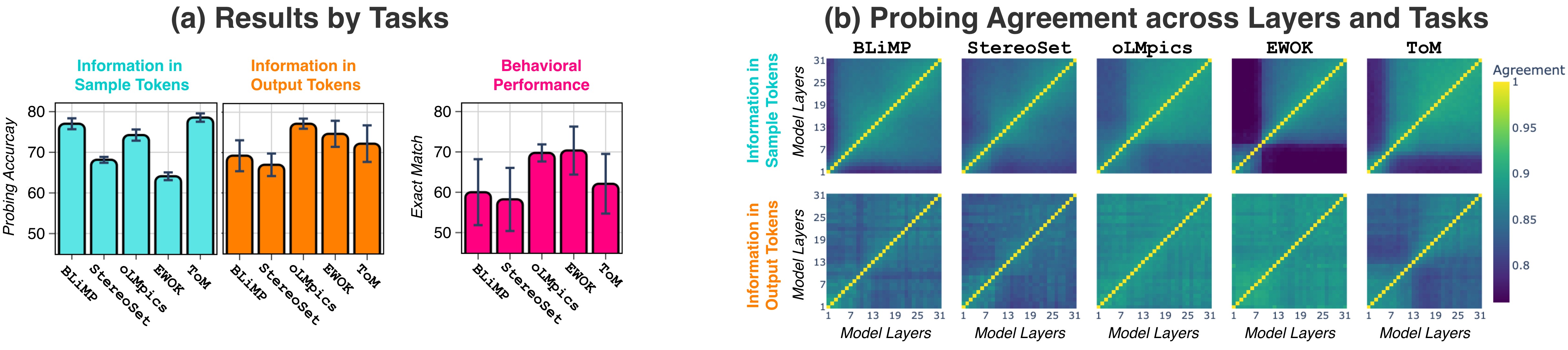}
    \caption{
       \textbf{(a)} Task-specific information in \tsample{} tokens, \toutput{} tokens, and behavioral performance (EM) across judgment tasks, averaged across models and prompting variations.
        \textbf{(b)} Layer-wise pairwise representation agreement heatmaps per task, for \tsample{} tokens (top) and \toutput{} tokens (bottom).
        Each cell $(i,j)$ indicates mean agreement between probing predictions at layers $i$ and $j$, averaged across instances.
     }
    \label{fig:task_results}
\end{figure*}

\paragraph{Behavior consistently aligns with production.}
Task-specific information in sample tokens varies across tasks (\autoref{fig:task_results} (a)), with higher values for \blimp{} and \tom{} and lower values for \ewok{}, but does not follow task categories---even closely related tasks such as \olmpics{} and \ewok{} differ.
In contrast, production shows more consistent structure, with syntactic tasks (\blimp{}, \stereoset{}) differing from knowledge and reasoning tasks (\olmpics{}, \ewok{}, \tom{}).
Task-specific information across stages is uncorrelated ($\tau=0$), indicating that processing-stage information does not predict production-stage information.
Instance-level Kendall $\tau$ correlations show that output token probing more reliably predicts behavior than sample token probing across all tasks (\ewok{}: $\tau=0.70$ vs.\ $0.29$; \stereoset{}: $0.58$ vs.\ $0.29$; \blimp{}: $0.56$ vs.\ $0.37$; \olmpics{}: $0.53$ vs.\ $0.21$; \tom{}: $0.22$ vs.\ $0.15$), with the gap varying systematically.

\paragraph{The processing--production coupling varies across tasks.}
\blimp{} shows the smallest gap, consistent with tight coupling and strong sensitivity to instruction removal (\autoref{sec:appendix_tasks}).
\stereoset{} occupies an intermediate position.
\olmpics{} and \ewok{} show the largest gaps, indicating that behavior is largely determined by production regardless of processing-stage information, and both remain robust when instructions are removed (\autoref{sec:appendix_tasks}).
\tom{} is a special case: both correlations are weak ($\tau=0.22$ vs.\ $0.15$), and removing instruction flow improves behavior, suggesting that instructions may interfere with processing (\autoref{sec:appendix_tasks}).
Overall, coupling is tighter for syntactic tasks, where surface form more directly constrains both stages.

\paragraph{Processing and production show distinct layer-wise dynamics.}
The layer-wise agreement heatmaps (\autoref{fig:task_results} (b)) provide a mechanistic perspective on the processing--production separation.
For sample tokens, low cross-layer agreement indicates that representations are continuously transformed across depth, with clear task-specific patterns: \ewok{} shows a sharp discontinuity around layer 10, suggesting a mid-network reorganization; \tom{} exhibits isolated early-layer representations; and the remaining tasks show smoother transitions across layers.
In contrast, output tokens exhibit substantially higher and more uniform cross-layer agreement, indicating that production-stage representations stabilize early and are maintained throughout the network.
This difference links directly to behavior.
Tasks where output probing most reliably predicts behavior (\ewok{}, \olmpics{}) also show the broadest cross-layer agreement, suggesting that earlier and more stable production-stage commitment is associated with more reliable behavioral outcomes.
Notably, the task-specific structure observed in sample tokens does not carry over to output tokens, reinforcing that the two stages operate largely independently.

\paragraph{Summary}
Task type shapes the processing--production asymmetry along a spectrum from tight to loose coupling.
Across tasks, behavior consistently aligns with production, but the strength of this alignment depends on how strongly the task constrains the mapping from internal representations to outputs.
Thus, task differences primarily reflect how information is expressed, not whether it is encoded.

\section{Related Work}\label{sec:related_work}

\paragraph{Behavioral Effects of Instructions.}
Instruction tuning \citep{wei2021finetuned, Ouyang2022TrainingLM} substantially improves zero-shot generalization.
\citet{zhou2023lima} argue through the \textit{Superficial Alignment Hypothesis} that it primarily shapes how models express knowledge rather than what they know---a claim complemented by \citet{min2022rethinking}, who show that randomly replacing demonstration labels barely hurts in-context learning performance, suggesting that models rely more on structural prompt properties than on label semantics.
Models also show strong sensitivity to surface formats \citep{ashury2025mighty,DBLP:conf/iclr/Sclar0TS24, mizrahi-etal-2024-state,ashury2026robustness}, few-shot example ordering \citep{lu2022fantastically}, and output distribution biases such as surface-form competition \citep{holtzman-etal-2021-surface} and recency or majority-label effects \citep{zhao-etal-2021-calibrate}.
Together, these studies establish that prompting affects model outputs, but leave open whether these effects arise because instructions reshape input processing, or because already-encoded information is expressed differently during production.

\paragraph{What Models Encode vs.\ What They Express.}
Probing classifiers \citep{belinkov-2022-probing, tenney2019you, conneau-etal-2018-cram, hewitt-liang-2019-designing, voita-titov-2020-information} reveal a consistent pattern: internally encoded knowledge does not always surface in model outputs.
Models encode more than they express \citep{burns2023discovering, gekhman2025insideout, orgad2025llms, feng2025monitoring, azaria-mitchell-2023-internal, slobodkin-etal-2023-curious}, and behavioral evaluations therefore underestimate their capabilities \citep{hu2024auxiliary, heo2025llmsknow, tutek-etal-2025-measuring}.
Mechanistic analyses narrow this gap further, localizing task-following computations to specific attention heads and mid-layer directions \citep{todd2024function, olsson2022induction, hendel2023task, gottesman-geva-2024-estimating, meng2022locating}.
Most closely related, \citet{waldis2025aligned} jointly analyze behavior and internal representations for toxicity---showing that models encode more about input toxicity than their outputs reveal---and \citet{lepori2026representations} show that models struggle to deploy representations learned in-context for downstream predictions.
While these studies characterize the divergence between internal and behavioral perspectives, they do not study the impact of instructions on it, nor do they ground the input--output distinction in a theoretical framework that anticipates where asymmetries should arise.
In contrast, we offer a principled lens grounded in the cognitive processing--production distinction, operationalized via token positions, and show with correlational and causal evidence that instructions primarily act at output-token positions---filtering and transforming already-encoded information rather than reshaping what is encoded from the input.

\section{Discussion}\label{sec:discussion}
Results from more than half a million probing runs across \autoref{sec:results} to \autoref{sec:result_tasks} consistently reveal a \textbf{production-centered mechanism}: instructions act less as a gate on \textit{what} is encoded from the input and more as a filter that selects and transforms already-encoded information during the production of output tokens.
This contrasts with humans, where instructions influence both stages, including selective attention during processing.
The pattern holds across model families, scales, training stages, and tasks, varying in strength but not in kind.
The processing--production distinction, therefore, offers a principled analytical lens for studying LMs, even if LMs do not implement it the same way humans do.

\paragraph{Evaluation Implications}
Our findings directly impact how we evaluate LMs, as behavioral assessments conflate two failure modes: task-specific information may be absent during processing, or present but not selected during production.
Our results indicate that gains in more capable models are more driven by the latter, so behavioral evaluations not only underestimate internal capabilities \citep{slobodkin-etal-2023-curious,hu2024auxiliary,gekhman2025insideout,orgad2025llms} but can misattribute the origin of failures or improvements.
By the same logic, the known prompt sensitivity \citep{DBLP:conf/iclr/Sclar0TS24,mizrahi-etal-2024-state,habba-etal-2025-dove} is better understood as a production-stage phenomenon than as processing fragility.
Thus, evaluation research should generally distinguish between processing and production stages, as we should address these two failure modes differently: processing failures require strengthening task-relevant encoding during training, whereas production failures can be addressed at generation time through output calibration \citep{zhao-etal-2021-calibrate}, inference-time steering \citep{li-etal-2024-inference}, richer contextual grounding \citep{zhuo-etal-2024-prosa}, or inference scaling with additional \textit{thinking} tokens \citep{DBLP:journals/corr/abs-2501-12948}.

\paragraph{Scale and Alignment as Production-Stage Phenomena}
The asymmetry also reframes two well-studied axes of model improvement.
Scaling model size yields disproportionate gains at output token positions (\autoref{sec:result_models}), suggesting that the behavioral benefits of scaling come disproportionately from better \textit{expression} of already-encoded information, not from encoding more of it.
Instruction-tuning shows the same signature more sharply: instruction-tuned models carry substantially more task-specific information at output token positions than their base counterparts, while sample token representations remain comparably unchanged.
This gives a direct mechanistic reading of the \textit{Superficial Alignment Hypothesis} \citep{zhou2023lima}: post-training need not---and in our measurements, largely does not---restructure what the model encodes from its inputs, and instead reshapes how that information is gated into outputs.

\paragraph{Efficiency and Model Development}
The asymmetry also opens concrete directions for efficiency and model development.
If instruction tokens have minimal impact on sample encoding, key-value caches could de-prioritize instruction-token representations during sample processing with limited loss in task-relevant information.
The relative stability of sample token representations is also consistent with assumptions made by neurosymbolic approaches \citep{garcez-lamb-2023} and shows potential for grounding symbolic operations.
Finally, the asymmetry raises questions for pre-training and alignment objectives. 
The pattern may reflect that instruction-following is concentrated in the final stages of training, suggesting that encouraging instruction sensitivity during processing could reduce the asymmetry and bring LMs closer to the symmetric pattern observed in humans.
Whether such cognitive-inspired training would actually improve models, or whether the asymmetry is a natural outcome of training that enables current capabilities, is an open question for which our work offers a principled lens.

\section{Conclusion}
We investigated where instructions take effect in language models by analyzing internal representations at sample and output token positions, which operationalize the cognitive processing--production stages within decoder-only language models.
Two lines of future work follow from the production-centered mechanism we reveal.
First, whether and how this mechanism holds beyond binary-judgment tasks, such as open-ended generation, is an open empirical question.
Second, the asymmetry itself may reflect the training process of LMs, in which instruction-following is primarily emphasized in the final stages.
On this reading, our results point to potential gains from reducing the asymmetry and balancing the impact of instructions on both processing and production.
However, it remains an open question whether such cognitive-inspired training would actually improve models, or whether the asymmetry is a natural outcome of LM training that itself enables their current capabilities.
Our token-position operationalization provides a means for such future work: a principled, position-based decomposition that supports empirical tracking of the processing--production asymmetry and, more broadly, structured assessment of information within internal representations.

\section*{Limitations}
\label{app:limitations}

\paragraph{Task scope.}
Our evidence is based on binary judgment tasks with single-token outputs, chosen so that probing and behavior share the same target and can be directly compared.
Whether the production-centered mechanism extends to open-ended generation, multi-step reasoning, or longer outputs---and if so, whether it takes the same form---is an open question and a natural direction for future work.

\paragraph{Token positions as a proxy for processing and production.}
We operationalize the cognitive processing--production distinction via token positions: $\color{color-sample}\vec{h}_S$ at sample tokens as a proxy for the stage of language processing, $\color{color-output}\vec{h}_O$ at output tokens as a proxy for language production.
This is an approximation, not a claim of strict computational separation: upper-layer sample-token representations no longer reflect input encoding alone but incorporate task-relevant transformations that prepare the output, so $\vec{h}_S$ does not isolate the processing stage in a narrow encoding-only sense.
Conversely, early-layer output-token representations are still being transformed and not yet committed to generation, so $\vec{h}_O$ captures all computation over output tokens rather than the production stage in a narrow decoding-only sense.
Given the lack of sharp computational boundaries, we use this token-position abstraction as an analytical lens to locate instruction effects within the forward pass.
Even so, our results show that the two stages behave systematically differently across model families, scales, and tasks, indicating that the token-position distinction tracks a meaningful computational separation despite the absence of architectural boundaries in decoder-only models.

\paragraph{Linear decodability.}
Linear probes provide a lower bound on the linearly accessible information, so the stability we observe in sample representations could, in principle, partly reflect the limits of linear decodability rather than the representation's full information content.
Our selectivity \citep{hewitt-liang-2019-designing}, information-theoretic \citep{voita-titov-2020-information}, and non-linearity checks (Appendix \autoref{sec:probing_reliability}) directly address this: probes show high selectivity against control tasks, information-theoretic and accuracy-based measurements reveal consistent patterns, and adding non-linear capacity to the probes does not change the layer-wise information dynamics we report.
Together, these validations robustly demonstrate that the observed asymmetry reflects representational structure rather than a probing artifact, although any model interpretability method can only approximate internal representations in the absence of ground truth.

\paragraph{Confounds in the attention-blocking interventions.}
The $-58.0$~pp behavioral drop under the \texttt{full} intervention (\autoref{sec:results}) reflects at least two effects: a degraded ability to express task-relevant information in output tokens---confirmed by the accompanying drop in output-token probe accuracy---and the potential loss of output-formatting cues the model relies on to produce a well-formed binary answer.
The latter is a confound that the present design cannot fully disentangle.
We therefore read the behavioral drop as an upper bound on the task-information-specific contribution of the intervention---and as a concrete instance of the more general point made in \autoref{sec:discussion}: behavioral measurements alone cannot distinguish processing-stage from production-stage contributions.

\section*{Acknowledgments}
We thank Polina Tsvilodub and Michael Franke for their valuable feedback and discussions.
Andreas Waldis is supported by the Volkswagen Foundation through a Momentum grant, by the state of Baden-Württemberg through bwHPC, and the data storage service SDS@hd, supported by the Ministry of Science, Research, and the Arts Baden-Württemberg (MWK).

\bibliography{tmlr,anthology}
\bibliographystyle{tmlr}
\appendix
\section{Appendix}
\subsection{Model Checkpoints}
\label{app:checkpoints}

\autoref{tab:checkpoints} lists the Hugging Face checkpoints of the 14 individual evaluated models used throughout this work. 
For each family, we use both the base and instruction-tuned variants where available.

\begin{table}[h]
\centering
\small
\begin{tabular}{lll}
\toprule
Family & Main experiments & Scaling analysis (sizes) \\
\midrule
Llama-3.1 & \texttt{meta-llama/Llama-3.1-8B(-Instruct)} & 8B \\
OLMo-2    & \texttt{allenai/OLMo-2-1124-7B(-Instruct)} & 1B, 7B, 13B, 32B \\
Qwen-2.5  & \texttt{Qwen/Qwen2.5-7B(-Instruct)}        & 0.5B, 1.5B, 3B, 7B, 14B, 32B \\
\bottomrule
\end{tabular}
\caption{Checkpoints used throughout the paper. Base and instruct variants are evaluated in parallel where available.}
\label{tab:checkpoints}
\end{table}

\subsection{The Reliability of the Probing Setup}\label{sec:probing_reliability}

\begin{figure*}[h]
    \centering
    \includegraphics[width=0.99\textwidth]{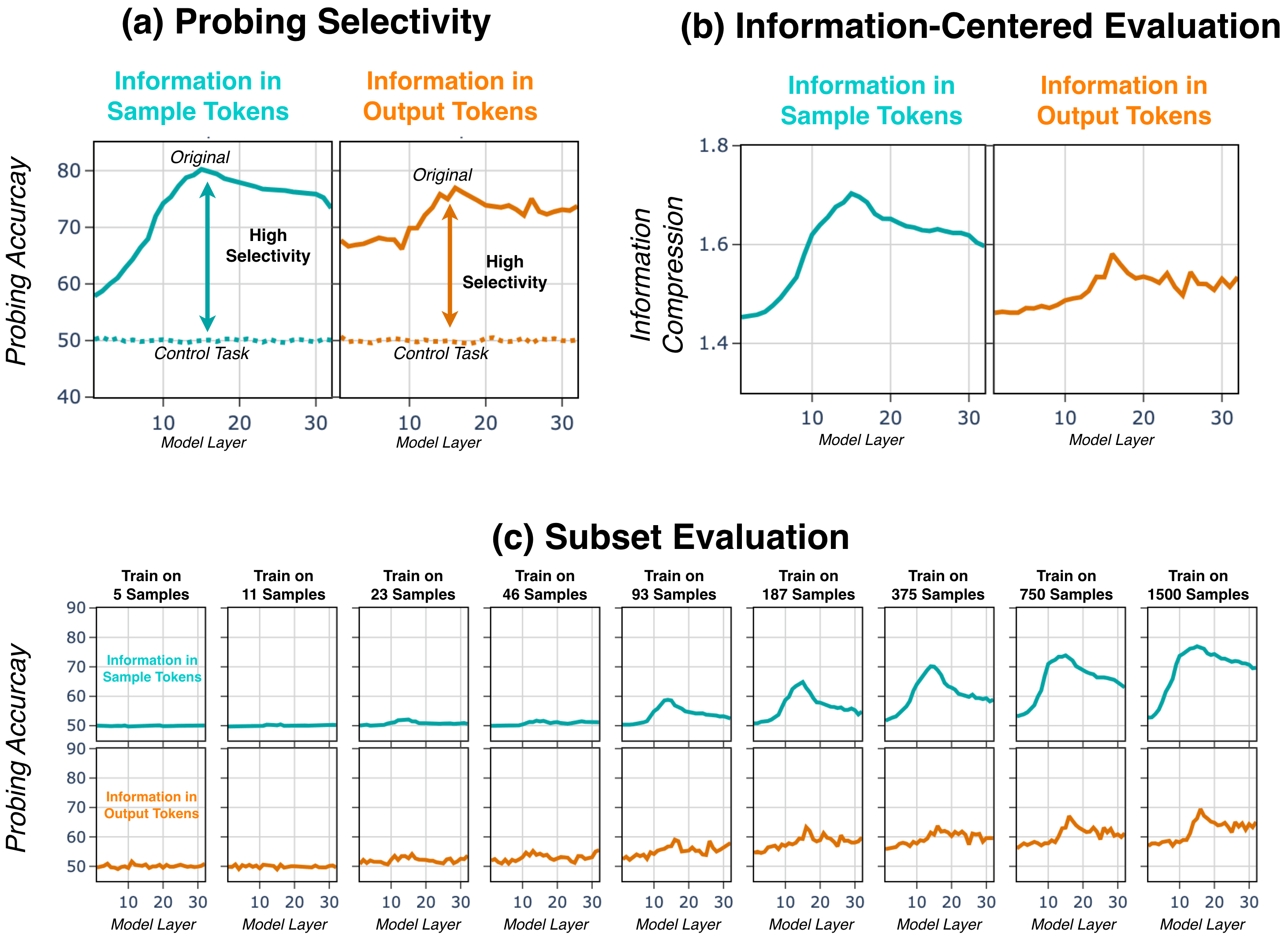}
    \caption{
        Validation of the probing setup across model layers, averaged across tasks and models.
        \textbf{(a)} Probing selectivity: accuracy of probes trained on original representations compared to control tasks \citep{hewitt-liang-2019-designing}, shown separately for \tsample{} and \toutput{} tokens.
        High selectivity confirms that probes capture task-relevant structure rather than exploiting their own capacity.
        \textbf{(b)} Information-theoretic assessment \citep{voita-titov-2020-information} of task-specific information in \tsample{} and \toutput{} token representations, showing consistent patterns with accuracy-based probing results.
        \textbf{(c)} Probing accuracy as a function of training set size, confirming that stable information patterns emerge with as few as 100 to 200 samples.
    }
    \label{fig:probg_verify}
\end{figure*}

\begin{figure*}[h]
    \centering
    \includegraphics[width=0.99\textwidth]{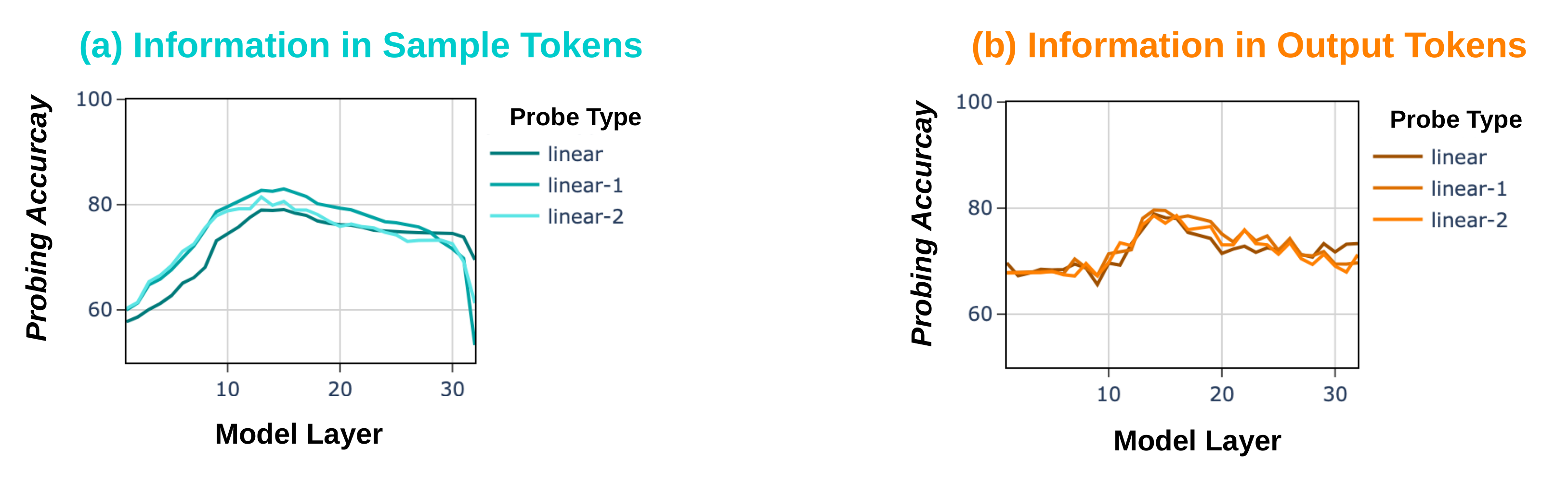}
    \caption{
        Validation of the probing setup with no intermediate hidden layer (\textit{linear}). 
        We also assess task-specific information with \tsample{} (a) and \toutput{} (b) tokens using a probe with either one (\textit{linear-1}) or two intermediate layers (\textit{linear-2}) of dimension 100.
        Results show slight variance in information strength for sample tokens with \textit{linear-1}, but no substantial change in the overall information pattern across model layers. 
    }
    \label{fig:linearity}
\end{figure*}

\autoref{fig:probg_verify} validates the probing setup along three dimensions across all model families, sizes, and tasks.
First, we report high probing selectivity using control tasks \citep{hewitt-liang-2019-designing}, confirming that probes learn from the representations rather than from their own capacity.
Second, we assess probes from an information-theoretic perspective \citep{voita-titov-2020-information}, and we find consistent patterns with the accuracy-based results reported in the main text.
Third, we verify that as few as 100 to 200 samples are sufficient to reveal stable information patterns, confirming that our findings are not an artifact of sample size.
Finally, we compare in \autoref{fig:linearity} the probing results with no intermediate layer (\textit{linear}) to those with one (\textit{linear-1}) or two (\textit{linear-2}) layers. 
Since nonlinear probes yield only slight differences in information levels without changing the overall layer-wise pattern, we conclude that linear probes capture task-specific information with sufficient reliability.

\subsection{Sanity Checks of the Impact of Instruction}\label{sec:task_demands}

We verify the observed variation in the impact of instructions on language processing and production using sanity prompts that introduce semantic noise or increase task difficulty, based on the best-performing variation ($\mathcal{P}_{\curvearrowleft}$).

\begin{figure*}[h]
    \centering
    \includegraphics[width=0.7\textwidth]{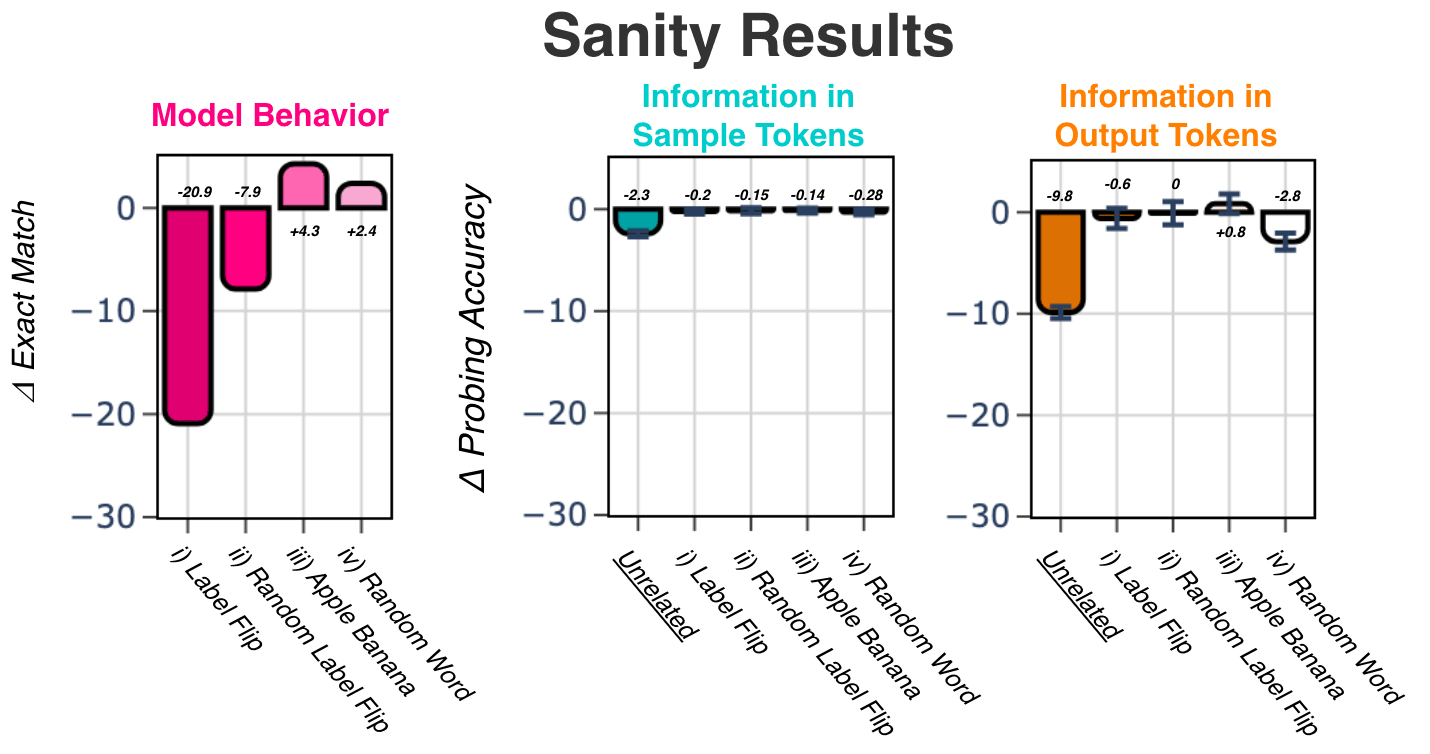}
    \caption{
        Sanity checks of the production-centered mechanism, averaged across tasks and models.
        \textbf{Left}: impact of semantically unrelated instructions on task-specific information in \tsample{} and \toutput{} tokens and on behavioral performance, relative to the standard instruction-first variation ($\mathcal{P}_{\curvearrowleft}$).
        \textbf{Right}: impact of increasing task demands---label flip, random label flip, abstract answers (apple/banana), and random word pairs---on behavior, \tsample{} token information, and \toutput{} token information.
        In both conditions, \tsample{} token information remains largely stable while behavioral performance and \toutput{} token information are substantially affected, confirming that the processing stage is robust to instruction variation.
    }
    \label{fig:sanity}
\end{figure*}

\paragraph{Unrelated instructions primarily change language production.}
We first verify our previous results by introducing semantically unrelated instructions that are irrelevant to the judgment task.
Specifically, we ask the LMs whether a given sentence contains the letter ``\textit{a}'' a specific number of times---expecting ``\textit{yes}'' when the specific number corresponds to the sentence and ``\textit{no}'' if not.
Following results shown in \autoref{fig:sanity}, these unrelated instructions caused a substantial drop ($-10.0$ probing accuracy) in task-specific information in the language production stage (output tokens).
In contrast, the information loss within sample tokens, representing the language processing stage, is much lower ($-2.0$).
These results confirm the primary mechanism: instructions primarily affect how information is transformed into output text during production, but not how information from the input is initially encoded.

\paragraph{Increasing task demands mainly impact model behavior.}
Second, we verify the mechanism by increasing task demands via additional required reasoning steps: \textit{i)} reversing the required label (``\textit{no}'' instead of ``\textit{yes}'' for positive judgments); \textit{ii)} randomly applying this label flip; \textit{iii)} requiring abstract answers (``\textit{apple}'' for positive, ``\textit{banana}'' for negative); and \textit{iv)} requiring random word pairs.
Behavioral results from \autoref{fig:sanity} align with prior work \citep{webson-pavlick-2022-prompt} and show a substantial impact of these variations, with up to $-20.9$ less performance when \textit{flipping} the label meaning for all samples.
Notably, we found only a minor impact on task-specific information of these prompting variations in both sample and output tokens.
Consistent with our previous findings, these insights suggest that instructions activate a shared, stable, and latent knowledge base but employ an unstable function to apply that knowledge.
The stability of task-specific information, particularly during input encoding, is maintained, only varying during language production when contrasting high-impact prompting variations (like $\mathcal{P}_{\curvearrowleft}$ vs. $\mathcal{P}_{\curvearrowright}$).
These results extend previous findings that behavioral assessments are unreliable \citep{hu2024auxiliary,DBLP:journals/corr/abs-2403-00998} and suggest that information-based measures offer a more robust view of underlying knowledge.

\subsection{Detailed Instance-Level Probing--Prompting Alignment}
\label{app:instance_alignment}

\begin{figure*}[h]
    \centering
    \includegraphics[width=0.99\textwidth]{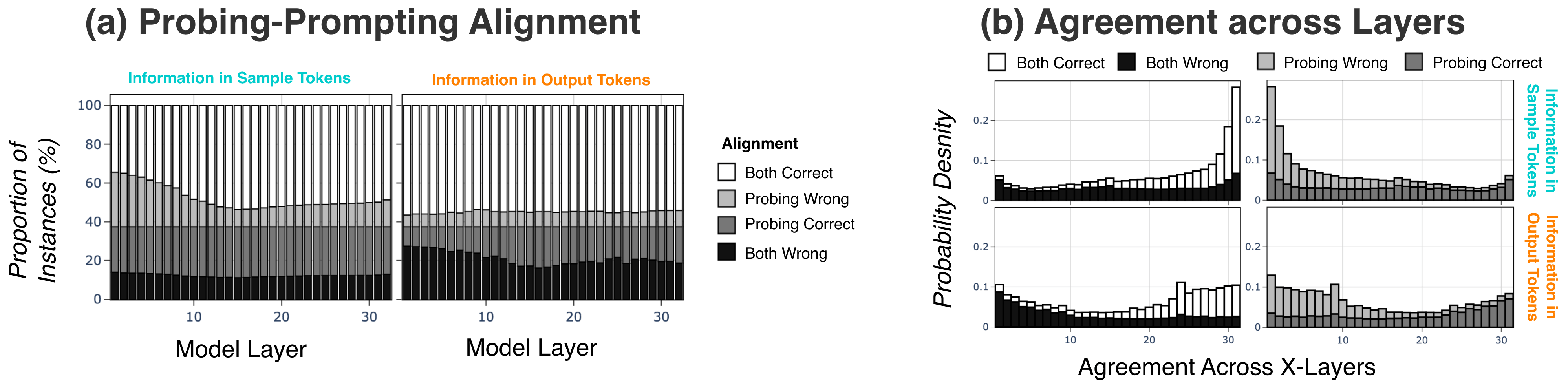}
    \caption{
        Instance-level probing--prompting alignment, averaged across tasks and models.
        \textbf{(a)} Proportion of instances per layer falling into each agreement category for \tsample{} (left) and \toutput{} (right) tokens: both correct (white), probing wrong only (light gray), probing correct only (dark gray), and both wrong (black).
        \textbf{(b)} Probability density of the number of layers across which each agreement category holds consistently, shown separately for \tsample{} and \toutput{} tokens.
        Correct agreement on \tsample{} tokens is concentrated at maximum layer consistency, reflecting stable processing-stage encoding, whereas correct agreement on \toutput{} tokens is more broadly distributed, reflecting greater fragility of production-stage alignment.
    }
    \label{fig:task_alignment}
\end{figure*}

\autoref{fig:task_alignment} shows the breakdown of correctness for probing--prompting alignment across model layers, extending the variation-level agreement analysis in \autoref{sec:results} by showing not only how often probing and prompting disagree, but also why and how persistently.
For sample tokens, the dominant disagreement is cases where probing is correct but prompting fails, which decreases across layers as both converge in upper layers.
This shows that processing-stage representations encode task-relevant information before the model expresses it behaviorally---probing detects this structure early, but the model has not yet translated it into correct output.
Importantly, when both are correct for sample tokens, this agreement holds consistently across nearly all layers, confirming that correct processing-stage encoding is robust and stable once established.
Disagreement categories, by contrast, tend to be layer-specific rather than persistent.
For output tokens, the dominant disagreement is cases where both probing and prompting fail, most pronounced in early layers and decreasing toward upper layers.
Unlike sample tokens, these errors persist across more layers, indicating that production failures reflect a more fundamental lack of reliable task-relevant structure rather than a transitional encoding gap.
Panel (b) further shows that ``both correct'' for output tokens is less consistently maintained across layers than for sample tokens, meaning that correct production-stage encoding is more layer-dependent and less stable.
Together, these results show that processing and production not only differ in their sensitivity to instructions but also in the stability of their encoding: processing develops task-relevant structure early and maintains it robustly, while production alignment is more fragile and depth-dependent.

\subsection{Absolute Layer Comparison Across Model Sizes}\label{app:scaling_absolute}

\begin{figure*}[h]
    \centering
    \includegraphics[width=0.99\textwidth]{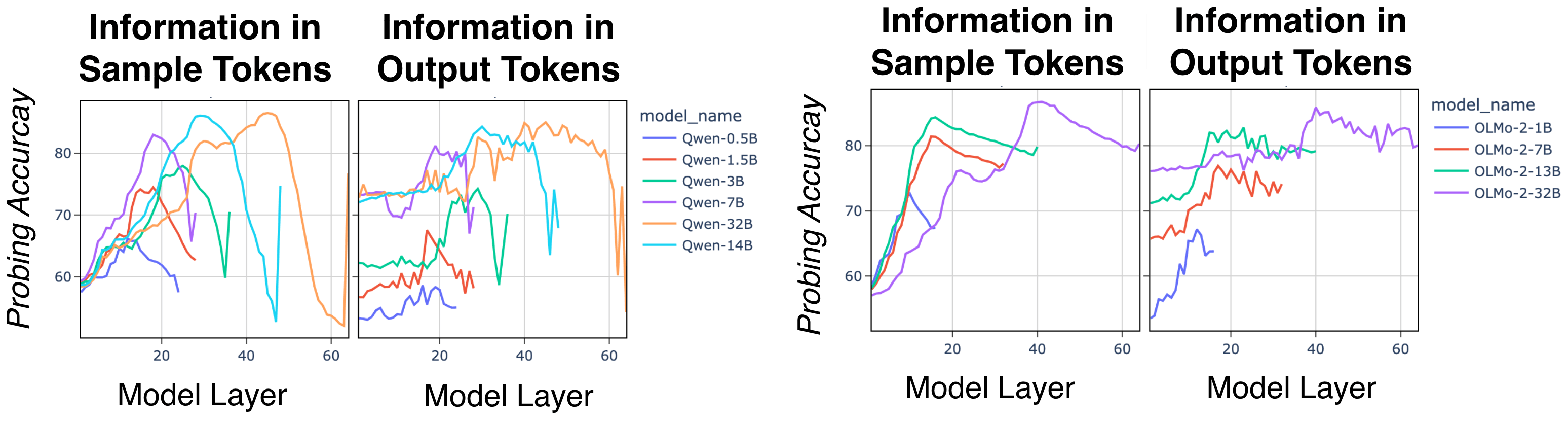}
    \caption{
        Layer-wise task-specific information for \tsample{} (left) and \toutput{} (right) tokens across five \qwen{} model sizes, shown in terms of absolute layer positions.
        Each curve represents one model size.
        For \tsample{} tokens, all sizes converge at similar early-layer accuracy and diverge only as they rise toward their peaks, suggesting that scaling extends processing-stage encoding depth rather than fundamentally changing how it starts.
        For \toutput{} tokens, smaller models (0.5B--3B) show flat, narrow curves while larger models (7B--32B) rise substantially higher across a broader layer range, reflecting the more qualitative shift in production-stage encoding with scale.
    }
    \label{fig:scaling_absolute}
\end{figure*}

While the main analysis focuses on relative layer positions to allow comparison across models of different sizes, comparing models in terms of absolute layer positions reveals two additional patterns, as shown in \autoref{fig:scaling_absolute}.
First, for sample tokens, all model sizes start at similar probing accuracy levels in the early layers, only diverging as they rise toward their respective peaks.
This suggests that the basic encoding of task-specific information during processing emerges similarly regardless of model size, and that scaling primarily extends how far and how long this encoding develops rather than changing how it starts.
Second, for output tokens, the picture is different.
Smaller models (0.5B, 1.5B, 3B) exhibit flat, narrow encoding curves throughout, whereas larger models (7B, 32B) rise substantially higher and retain task-specific information across a much broader range of layers.
Rather than a smooth continuation as in sample tokens, the production stage does not scale gradually but changes more fundamentally above a certain model size.
These observations suggest that while processing-stage encoding scales smoothly and continuously, production-stage encoding undergoes a more qualitative shift with model size---consistent with the main text's finding that scaling disproportionately benefits production.

\subsection{Detailed Task Results}
\label{sec:appendix_tasks}

\begin{figure*}[h]
    \centering
    \includegraphics[width=0.80\textwidth]{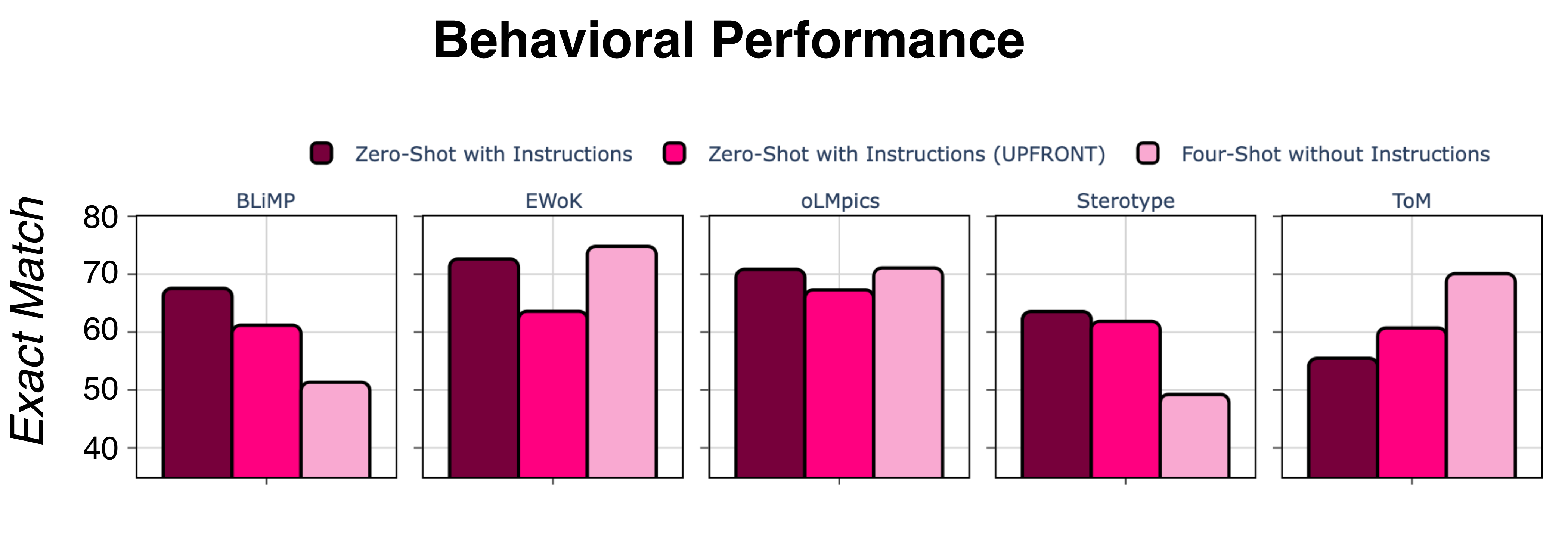}
    \caption{
        Behavioral performance (EM) across the three prompting variations ($\mathcal{P}_{\curvearrowleft}$, $\mathcal{P}_{\curvearrowright}$, $\mathcal{P}_{w/o}$) for each judgment task, averaged across models.
        \blimp{} shows the strongest sensitivity to instruction removal, while knowledge and reasoning tasks remain largely robust.
        \tom{} is the only task that benefits from few-shot demonstrations over explicit instructions.
    }
    \label{fig:task_behavioral}
\end{figure*}

\paragraph{Behavioral Results Across Prompting Variations}
\autoref{fig:task_behavioral} shows behavioral performance across prompting variations for each task.
The most pronounced variation arises from few-shot prompting without instructions ($\mathcal{P}_{w/o}$).
For \blimp{}, replacing instructions with four-shot examples substantially decreases performance ($\Delta \approx -15.0$), consistent with the strong sensitivity of surface-sensitive tasks to instruction-driven production established in \autoref{sec:result_tasks}.
In contrast, \tom{} benefits from concrete demonstrations, yielding the highest performance under $\mathcal{P}_{w/o}$, suggesting that examples are more informative than abstract task descriptions for reasoning about mental states.
Knowledge tasks (\ewok{}, \olmpics{}) remain largely robust across variations, consistent with their stable behavior--output coupling.
Instruction placement has comparatively minor effects overall, with a slight advantage for standard ordering on form-sensitive tasks (\blimp{}, \stereoset{}).
\stereoset{} drops noticeably under few-shot prompting, reflecting its particular sensitivity to stereotype-related content in demonstrations.

\begin{figure*}[h]
    \centering
    \includegraphics[width=0.80\textwidth]{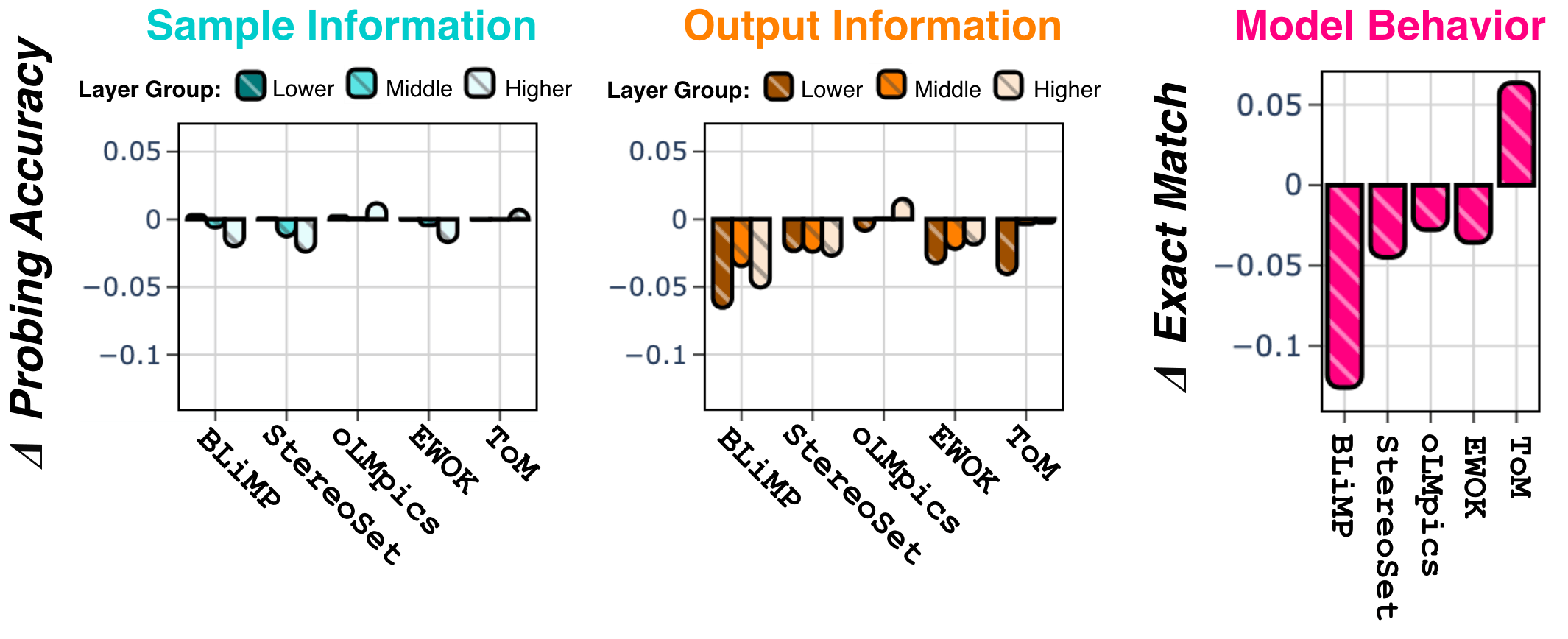}
    \caption{
        Effect of the prompt-only intervention on task-specific information in \tsample{} and \toutput{} tokens and on behavioral performance, shown per judgment task and averaged across models.
        Results are reported as differences relative to the unmodified instruction-first variation ($\mathcal{P}_{\curvearrowleft}$), grouped by lower, middle, and upper model layers.
        \tsample{} token information remains stable across all tasks, while \toutput{} token information and behavioral performance show task-specific responses that reflect the processing--production spectrum established in \autoref{sec:result_tasks}.
    }
    \label{fig:intervention_tasks}
\end{figure*}

\paragraph{Causal Intervention Results}
\autoref{fig:intervention_tasks} shows the effect of the prompt-only intervention across tasks, which cuts attention between instruction and sample tokens under the instruction-first prompting variation ($\mathcal{P}_{\curvearrowleft}$).
Across all tasks, sample token information remains largely stable, with only minor deviations visible in the upper model layers ($\leq -2.0$ for \blimp{} and \stereoset{}), confirming that the processing stage is largely unaffected by instruction flow regardless of task type.
The production stage and behavior show stronger and more task-specific responses.
\blimp{} exhibits the largest effects: output token information drops by $-4.0$ to $-7.0$ across layers, and behavioral performance falls by $-13.0$, consistent with the tighter processing--production coupling and greater sensitivity to instruction-driven production identified in \autoref{sec:result_tasks}.
Knowledge tasks (\olmpics{}, \ewok{}) show minimal behavioral effects in either direction, reflecting the stable behavior--output coupling and the decoupling of processing from behavior established in \autoref{sec:result_tasks}.
\tom{} yields a positive behavioral effect ($+6.0$), indicating that instruction tokens reaching sample tokens introduce interference for this reasoning task rather than providing a useful signal.
This further supports the view that for knowledge and reasoning tasks, the processing stage is not only decoupled from behavior but can actively work against it, making production the sole reliable pathway to correct output.

\begin{figure*}[h]
    \centering
    \includegraphics[width=0.99\textwidth]{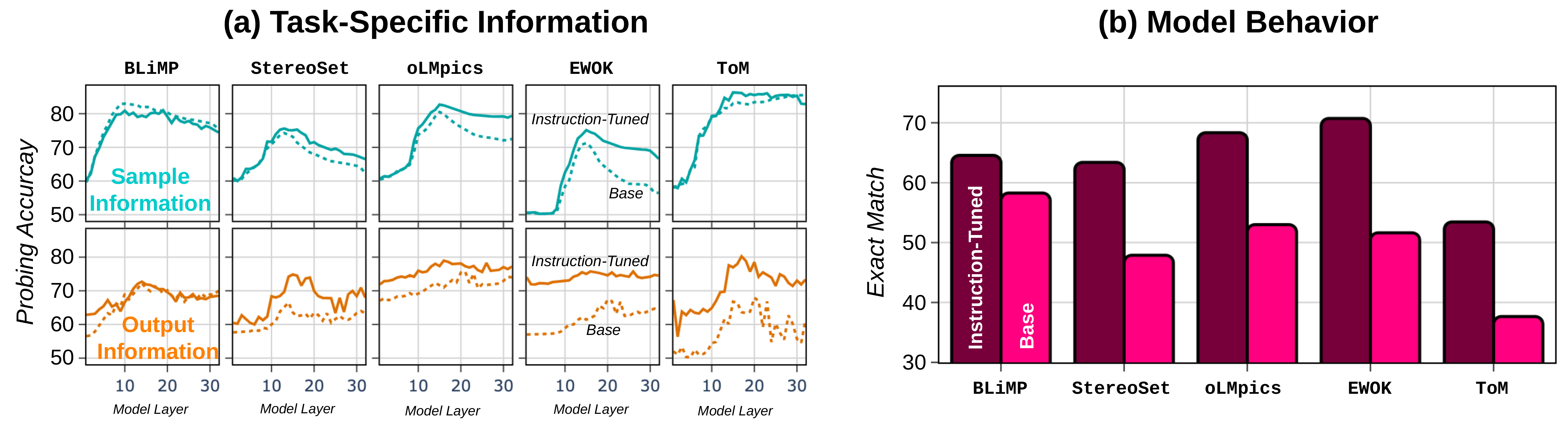}
    \caption{
        Comparison of pre-trained (\textit{base}, dotted) and instruction-tuned models (solid) per judgment task.
        \textbf{(a)} Layer-wise task-specific information in \tsample{} (top) and \toutput{} (bottom) tokens for each task.
        \tsample{} token curves are nearly identical across conditions for all tasks, while \toutput{} token curves show task-dependent gains from instruction-tuning, largest for knowledge and reasoning tasks.
        \textbf{(b)} Behavioral performance (EM) for base and instruction-tuned models per task.
        Instruction-tuning improves behavior most for tasks where production is the primary behavioral determinant, consistent with post-training operating primarily at the production stage.
    }
    \label{fig:base_chat_tasks}
\end{figure*}

\paragraph{Task-Level Impact of Instruction-Tuning}
\autoref{fig:base_chat_tasks} compares base and instruction-tuned models per task, extending the aggregated comparison in \autoref{sec:result_models}.
Across all tasks, sample token curves for base and instruction-tuned models overlap closely, confirming that post-training does not substantially change how task-relevant information is encoded during processing regardless of task type.
Instruction-tuning consistently increases output token information, but the magnitude of this gain varies with the task spectrum: the largest production-stage gains occur for knowledge and reasoning tasks (\olmpics{}, \ewok{}, \tom{}), where production is already the primary behavioral determinant, while gains are more modest for surface-sensitive tasks (\blimp{}, \stereoset{}).
For \stereoset{}, the behavioral gap between base and instruction-tuned models remains small despite the low behavioral ceiling, consistent with the production-stage disruption for this task being structural rather than a consequence of insufficient instruction-tuning.

\begin{figure*}[h]
    \centering
    \includegraphics[width=0.99\textwidth]{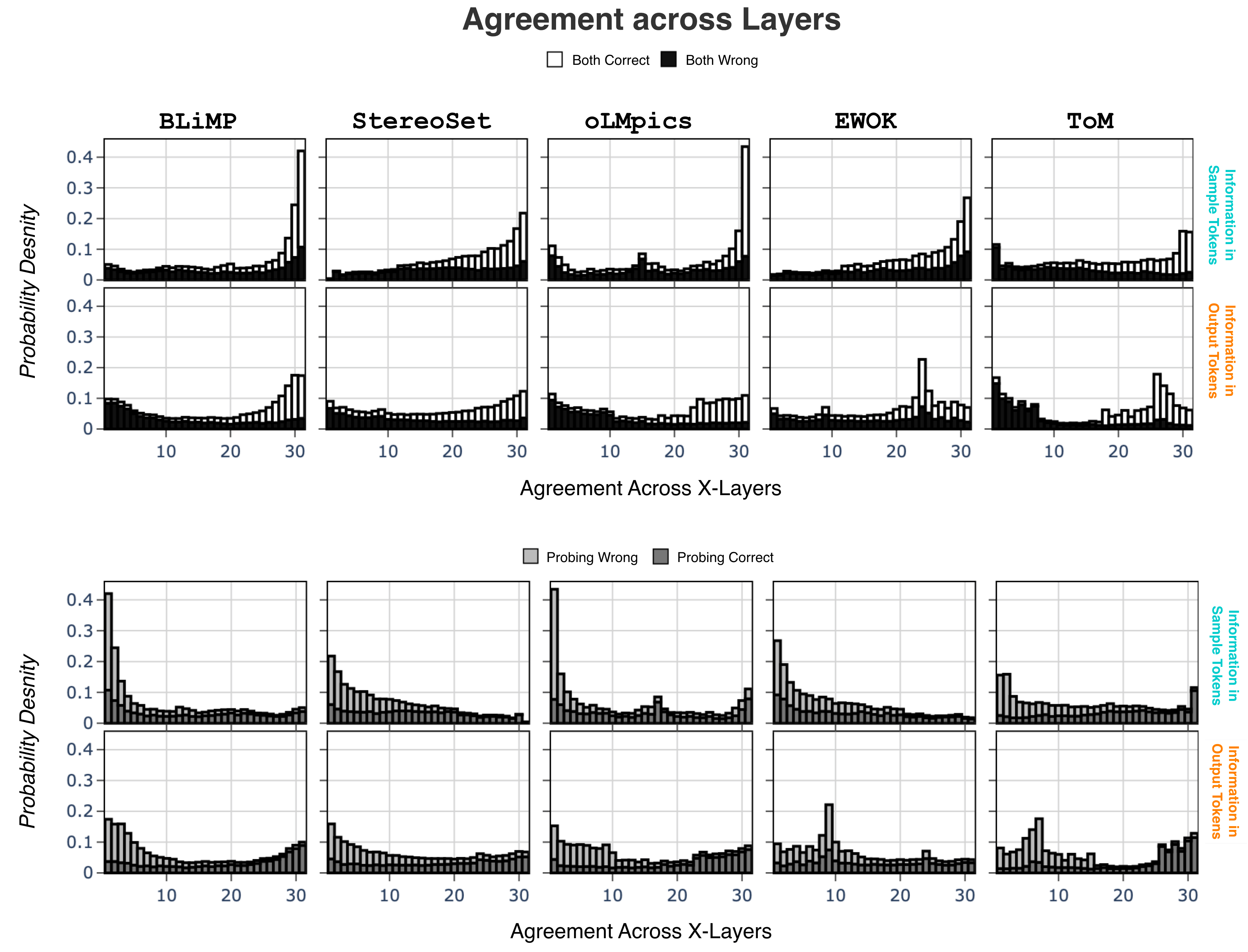}
    \caption{
        Layer-wise probing--prompting consistency distributions per judgment task, for \tsample{} (bottom rows) and \toutput{} (top rows) representations.
        The x-axis indicates the number of layers across which a given probing prediction is consistent.
        \textbf{Top panels}: cases where probing and prompting agree, split by whether the prediction is correct (white) or wrong (black).
        \textbf{Bottom panels}: cases where probing and prompting disagree, split by whether probing is correct (white) or wrong (black).
        For \blimp{}, correct output agreement is sharply concentrated at maximum consistency.
        For \stereoset{}, correct and wrong agreement distributions are comparably spread, reflecting unstable production-stage commitments.
        For knowledge and reasoning tasks, correct output agreement concentrates at maximum consistency while input distributions remain flat, illustrating the processing--production decoupling.
    }
    \label{fig:task_layer_agreement}
\end{figure*}

\paragraph{Task-Level Layer-Wise Agreement of Probing and Prompting}
\autoref{fig:task_layer_agreement} extends the aggregated consistency analysis in \autoref{app:instance_alignment} by showing the layer-wise agreement distributions per task, separately for cases where probing and prompting agree (top panels) and disagree (bottom panels).
For \blimp{}, correct agreement on output representations is sharply concentrated at maximum layer consistency, indicating that tight processing--production coupling translates into stable correct production.
For \stereoset{}, wrong and correct agreement on output representations are comparably distributed across consistency values, lacking the sharp concentration at maximum consistency seen for other tasks---consistent with the production-stage disruption identified in \autoref{sec:result_tasks}.
For knowledge and reasoning tasks (\olmpics{}, \ewok{}), correct output agreement is also concentrated at maximum consistency, reflecting reliable production once engaged, while input representations show no such concentration, confirming the decoupling of processing from behavior for these tasks.
\tom{} shows diffuse distributions across all quadrants, consistent with the interference introduced by instruction tokens at the processing stage identified in the intervention results above.

\end{document}